%% file: main.tex
\documentclass[10pt,a4paper,twocolumn]{article}

\usepackage[T1]{fontenc}
\usepackage[utf8]{inputenc}
\usepackage{lmodern}
\usepackage[english]{babel}

\usepackage[margin=1.9cm]{geometry}
\usepackage{amsmath,amssymb}
\usepackage{graphicx}
\graphicspath{{figures/}}
\usepackage{booktabs}
\usepackage{float}
\usepackage{dblfloatfix}
\usepackage{caption}
\usepackage[numbers,sort&compress]{natbib}
\usepackage{authblk}

\usepackage{xcolor}
\definecolor{linkcol}{RGB}{25,60,120}
\usepackage[colorlinks=true,allcolors=linkcol]{hyperref}
\usepackage{microtype}
\title{\bfseries From Foundation Embeddings to Cropland Maps:\\
Label Efficiency, Temporal Transferability and Independent Human Validation}

\author[1]{Mohammad Ammar Mughees}
\author[1]{Giovanni Montefoschi}
\author[2]{Zhongxin Chen}
\author[1]{Maria Antonia Brovelli}
\affil[1]{Department of Civil and Environmental Engineering, Politecnico di Milano, Milan, Italy}
\affil[2]{Food and Agriculture Organization of the United Nations, Rome, Italy}
\date{}

\newcommand{\Sph}{\mathcal{S}^{63}}

\newcommand{\rev}[1]{#1}

\begin{document}

\makeatletter
\twocolumn[
\begin{@twocolumnfalse}
\maketitle
\begin{center}
\begin{minipage}{0.94\textwidth}
\small
\noindent\textbf{Abstract.}
Geospatial foundation models provide reusable representations of satellite imagery that can
support downstream mapping with limited task-specific modelling. We evaluate whether annual
AlphaEarth embeddings support binary cultivated-versus-non-cultivated mapping in Maine, USA, using
192 spatially separated image patches and labels derived from the USDA Cropland Data Layer (CDL).
Without fine-tuning the foundation model, a lightweight classifier reaches approximately 93.7\%
overall accuracy and 90.8\% balanced accuracy on held-out patches. Logistic regression is within
0.3 percentage points of a gradient-boosted ensemble, while a nearest-class-centroid rule, which
uses labelled class centroids but no iterative parameter fitting, reaches 90.2\%. A balanced sample
of 60{,}000 labelled pixels is within approximately 1.3 percentage points of the full pool of
8.6 million pixels; because pixels are spatially autocorrelated, this result concerns pixel-sample
efficiency rather than 60{,}000 independent annotation sites. In a same-region temporal-transfer
experiment, classifiers trained in one year remain accurate across 2018 to 2023. Against a blind,
two-interpreter consensus at 385 randomly sampled points in one contiguous 2023 block, the
AlphaEarth-plus-random-forest map agrees at 95.3\% ($\kappa=0.82$), compared with 91.7\% for the CDL
($\kappa=0.72$; exact two-sided McNemar $p=0.0161$). This local result is consistent with partial
smoothing of CDL label noise, but it does not establish statewide correction of the reference
product. On the same points, the difference from a fine-tuned TerraMind segmentation model is not
statistically significant (95.3\% versus 93.5\%; $p=0.14$), and the experiment is not a controlled
comparison of computational cost. These results support frozen geospatial embeddings as a
low-compute candidate for regional cropland mapping, subject to the limits of a single-state study,
a 30\,m-derived training reference, and a one-block human validation.

\end{minipage}
\end{center}
\vspace{1.4em}
\end{@twocolumnfalse}
]
\makeatother

\input{sections/01_introduction}

\input{sections/02_data}

\input{sections/03_framework}

\input{sections/04_embedding_space}

\input{sections/05_classification_results}

\input{sections/06_label_efficiency}

\input{sections/07_cross_year}

\input{sections/08_human_validation}

\input{sections/09_discussion}

\input{sections/10_conclusions}

\section*{Data and Code Availability}
The full analysis pipeline is openly available under the MIT licence at
\url{https://github.com/Black-Lights/alphaearth-cropland-maine}. The repository contains the
\texttt{ae} Python package (data handling, sampling, models, ensembles, evaluation, mapping and the
embedding statistics), the notebooks that produce the figures and tables reported here, and the
wrappers that reproduce the classification and cross-year runs.

The AlphaEarth annual embeddings are the
\texttt{GOOGLE/SATELLITE\_EMBEDDING/V1/ANNUAL} collection in Google Earth Engine, and the labels are
derived from the USDA Cropland Data Layer (\texttt{USDA/NASS/CDL}) on the same platform by the
grouping given in Section~\ref{sec:cdlbinary}. The patches were cut with the open Earth Engine
toolkit of \citet{montefoschi2026gfm}, and the photo-interpretation was carried out in EarthLabel
\citep{earthlabel2026}. The 385-point consensus reference is available from the corresponding author
on reasonable request.

\section*{Funding}
\rev{This study was partially carried out within the Space It Up project funded by the Italian
Space Agency, ASI, and the Ministry of University and Research, MUR, under contract n.
2024-5-E.0 - CUP n. I53D24000060005.}

\section*{Acknowledgements}
The Maine patch dataset was prepared jointly by the first two authors, using the open Google Earth
Engine sampling toolkit of \citet{montefoschi2026gfm}, and the photo-interpretation reference was
produced by the two of them working independently and blind to the reference layer. The underlying
experimental work is documented in more detail in the first author's MSc thesis at Politecnico di
Milano \citep{mughees2026thesis}.

\section*{Disclaimer}
The views expressed in this paper are those of the authors and do not necessarily reflect the views
or policies of the Food and Agriculture Organization of the United Nations.

\small
\bibliography{refs}

\end{document}

%% file: sections/01_introduction.tex
\section{Introduction}
\label{sec:intro}

Producing land-use and land-cover information used to be slow and expensive. It relied on the
manual interpretation of imagery and on task-specific models that had to be built almost from
scratch for every new problem, and the biggest obstacle was data: training such models needed very
large, carefully labelled datasets, and high-quality labels are scarce because they are costly to
collect \citep{alphaearth2025}. Satellite missions such as Sentinel and Landsat have been both an
asset and a challenge in this respect. They provide a continuous stream of observations from which
a dataset can be built, but the volume of that data is too large to use directly, so a geospatial
expert has to spend considerable effort cleaning, compositing and engineering features before the
data becomes usable \citep{tessera2025}. Compositing, the usual way of dealing with clouds, also
discards part of the temporal phenological signal that matters most for crops.

Geospatial foundation models have begun to change this picture. Instead of engineering features by
hand, these models learn the features themselves, mostly through self-supervised pre-training, and
a single pre-trained model can then serve several downstream tasks. They divide into two broad ways
of working. One route provides a pre-trained backbone that the user still has to fine-tune for each
task, as with Prithvi \citep{jakubik2023foundation} or TerraMind \citep{terramind2025}. The other
route runs the model once, at the developer's side, and releases its output directly as
\emph{embeddings}: fixed-length per-pixel vectors that behave, in geospatial terms, like bands.
AlphaEarth \citep{alphaearth2025} and TESSERA \citep{tessera2025} follow this second route,
releasing 64-dimensional and 128-dimensional annual embeddings respectively. For every
$10\,\mathrm{m}\times10\,\mathrm{m}$ pixel the user obtains one such vector, which compactly
summarises a whole year of multimodal observations at that location, and which can be fed straight
into a light classifier with no heavy model to train or run.

This paper is concerned with that second family, and specifically with what it delivers for
cropland mapping. Four related terms are used here in a specific sense. A \emph{crop} is a
cultivated plant grown for food, fibre or fodder. \emph{Cropland} is land used to grow such crops,
comprising arable land together with land under permanent crops \citep{faostat_landuse}.
\emph{Cultivated land} is the broader category of land worked for agriculture; in the USDA Cropland
Data Layer it is represented by a dedicated Cultivated Layer, defined operationally as land cropped
in at least two of the previous five years \citep{boryan2011cdl}. \emph{Non-cultivated land} is
everything outside that category, including forest, water, developed, barren and similar
non-agricultural cover. The task addressed here is the binary separation of cultivated from
non-cultivated land, and, as explained in Section~\ref{sec:data}, we derive our own binary mask
from the annual 254-class Cropland layer rather than using the ready-made Cultivated Layer.

The premise that precomputed embeddings plus a simple classifier are sufficient has so far been
tested mostly by the model authors themselves. TESSERA reports that its embeddings, with only a
handful of labels, match or outperform task-specific models across five very different tasks and
without any fine-tuning \citep{tessera2025}. Independent evaluations are still few. The clearest we
found applies the same AlphaEarth and TESSERA embeddings to riverine habitat and geomorphic classes
with ordinary supervised classifiers, and reports a high potential for the approach
\citep{betz2026riverine}. Closer to our own task, a recent study evaluates AlphaEarth embeddings for
irrigated-cropland mapping in China and the United States, training random forests directly on the
64-dimensional vectors, and reports overall accuracies around 95 per cent with stable transfer
between years but weaker generalisation across regions \citep{yang2026irrigated}. Independent
evidence has accumulated quickly since. \citet{zvonkov2025cropland} map cropland in Togo from
precomputed Presto and AlphaEarth embeddings with a light random forest, reporting overall
accuracies of $0.897$ and $0.859$ respectively against a national test set. AlphaEarth is the weaker
of their two embeddings there, level with the GLAD product ($0.859$) and below WorldCover ($0.880$),
though the two embeddings are not evaluated on equal terms: their Presto embeddings span the 2019 to
2020 season of the reference labels, whereas the earliest AlphaEarth year available to them was
2021. \citet{ma2026harvesting} benchmark AlphaEarth embeddings
across three agricultural downstream tasks in the United States, crop-yield prediction, tillage
mapping and cover-crop mapping, and find them competitive with purpose-built remote-sensing models
when trained on local data, while reporting limited spatial transferability, low interpretability
and limited time sensitivity. \citet{lisaius2026senegal} apply the same paradigm to crop-type
classification under smallholder conditions in the groundnut basin of Senegal. Two further studies
examine the embedding space itself rather than a mapping task: \citet{benavides2026alphaearth}
characterise the functional roles of individual dimensions, and \citet{rahman2026geometry} the
geometry of the embedding manifold over the conterminous United States. We return to both in
Section~\ref{sec:embspace}. The agricultural
literature otherwise remains dominated by the fine-tuning and dedicated-training route: AgriFM is
pre-trained on more than 25 million multi-source samples and outperforms general-purpose foundation
models on crop mapping and field-boundary delineation \citep{li2026agrifm}, while fine-tuning
studies find that large models pull ahead of traditional machine learning mainly when labels are
scarce \citep{bourriz2026hyperspectral}. Both of those routes still have to train or fine-tune a
heavy model and attach a task head, which is precisely the cost the embedding paradigm avoids. A
broader benchmark of eight satellite foundation models is a useful warning in this context: it
finds performance strongly task-dependent, the quality of the pre-training data more important than
its sheer size, and the step from a ViT-Base to a ViT-Large backbone worth less than one per cent of
accuracy \citep{rotich2025robustness}. A larger model is not necessarily a better one, which is part
of why a simple classifier on a good embedding merits serious consideration.

The gap we address is therefore not whether embeddings can map cropland at all, which
\citet{zvonkov2025cropland} have now demonstrated, but how far such a map can be trusted. The
existing evaluations are reported as accuracy against a single reference product, on one region and
one year, and none of them examines the structure of the embedding space that produces the result,
tests whether a model fitted in one year still holds in another, or asks whether the reference
product itself is right where the map and the reference disagree. Whether a lightweight classifier
on these embeddings is competitive for binary cultivated mapping, how much labelled data it needs,
which properties of the embedding space decide the outcome, and how the resulting map compares with
independent human judgement rather than with the labels it was trained on, have not been examined
together for a single task.

In this paper we address that question over the state of Maine, in the United States, using the
annual AlphaEarth embeddings as the only input and labels derived from the USDA Cropland Data
Layer. We report four things. First, a statistical characterisation of the embedding space itself,
on the unit hypersphere where the vectors lie, which establishes which analyses are legitimate and
gives a classifier-free read on how hard the task is (Section~\ref{sec:embspace}). Second, the
accuracy of a graded set of light classifiers placed on the frozen embeddings, from a single linear
model to gradient-boosted ensembles (Section~\ref{sec:clfresults}). Third, how much labelled data
the approach actually needs, and how well a single-year model transfers across the years 2018 to
2023 (Sections~\ref{sec:labeleff} and~\ref{sec:crossyear}). Fourth, a validation that steps outside
the reference layer altogether, comparing the model, the Cropland Data Layer and a fine-tuned
segmentation model against independent human photo-interpretation on a common set of reference
points (Section~\ref{sec:humanval}). Throughout, we keep the classifier deliberately simple, so
that the embedding, and not the model placed on top of it, is the object of study.

Summarising, before entering into the details of our work, we want to emphasise the boundaries
we are considering. ``Label efficiency'' refers to the number of CDL-labelled pixels used by the
downstream classifier, not to the number of independent fields or manually annotated sites.
``Transferability'' refers to temporal reuse across years within the same Maine sampling frame;
geographic transfer to other regions is not tested. The human-reference comparison is point-based
and local to one contiguous block in 2023.

%% file: sections/02_data.tex
\section{Data and Study Area}
\label{sec:data}

\subsection{Maine as a study area}
The experimental work is carried out over the state of Maine, in the north-eastern United States.
Two practical constraints fixed this choice. The first is the ground truth: our labels are taken
from the USDA Cropland Data Layer, which is produced only for the United States, so the study area
had to lie within it. The second is the availability of the embeddings. AlphaEarth provides annual
embeddings globally, but we also wanted an area covered by the other precomputed-embedding model we
considered, TESSERA, whose coverage spans a more limited set of years and locations, so that the
same dataset could in principle be reused for a direct comparison in future work. Maine met both
conditions.

Maine is, however, a demanding area for an agricultural study, and this shaped the rest of the data
preparation. The state is dominated by forest, and cultivated land makes up only a small fraction of
its surface, concentrated in a few regions such as the potato land of the north and the blueberry
barrens and hay or dairy land elsewhere. Grouping the Cropland Data Layer into our binary target,
cultivated land accounts for only about 3\% of Maine in 2023, so it is a clear minority class, and
this becomes the central difficulty in building a usable dataset. An initial attempt confirmed how
severe the imbalance is: a single contiguous region of about $50\,\mathrm{km}$ by $50\,\mathrm{km}$
in central Maine turned out to be roughly 99.9\% non-cultivated. A single block of that kind does
not contain enough cultivated land in Maine to pose a balanced classification problem, and this is
what led us away from one large region and towards the patch-based, class-balanced sampling design
described in Section~\ref{sec:sampling}.

\subsection{Ground truth: from 254 classes to a binary mask}
\label{sec:cdlbinary}
Our labels come from the USDA Cropland Data Layer (CDL), produced annually by the National
Agricultural Statistics Service. The CDL is a raster land-cover map of the conterminous United
States in which every pixel is assigned one of 254 detailed classes, most of them specific crops. It
is generated by a supervised decision-tree classifier trained on ground truth from the Farm Service
Agency Common Land Unit programme, that is, the farmers' own field reports, together with satellite
imagery, and it is the standard crop-specific reference for the United States
\citep{boryan2011cdl}. For the years we use it was distributed at $30\,\mathrm{m}$ resolution. Its
accuracy is high for the major commodity crops, commonly around 85 to 95 per cent producer's and
user's accuracy, whereas rarer and more fragmented classes are less reliable \citep{boryan2011cdl},
and the layer is known to contain noisy pixels \citep{zhang2023cdlsam}. Since the CDL is the output
of a classifier rather than direct ground truth, we treat it as a reference to be tested rather than
as truth, which is the purpose of Section~\ref{sec:humanval}.

The CDL is distributed as two related products: the \textbf{annual Cropland layer}, the 254-class
map described above, which represents a single calendar year; and the \textbf{Cultivated layer}, a
ready-made binary map that NASS derives from several consecutive Cropland layers, labelling a pixel
as cultivated when it was cropped in at least two of the previous five years \citep{boryan2011cdl}.
We build our own labels from the annual Cropland layer rather than from the ready-made Cultivated
layer, and the reason is temporal. The Cultivated layer is defined over several years, which makes
it a single, slowly changing product that is almost identical from one year to the next and does not
represent the state of any individual year. Our study uses one embedding per year and examines how
the classification behaves across years, so we need a label that tracks each year on its own. Only
the annual Cropland layer provides this.

The grouping itself is straightforward. We assign to the cultivated class all of the genuine crop
classes, that is the field crops, the forage crops such as hay and alfalfa, and the tree and
specialty crops (CDL codes 1 to 60, 66 to 80, and 196 to 255). Everything else becomes
non-cultivated: pasture and grassland, fallow land, forest, shrubland, barren ground, developed
land, open water, and wetlands. The mask is built separately for each year of the study period,
2018 to 2023, so that each label is temporally consistent with the embedding it is paired with.

\begin{figure*}[tb]
\centering
\includegraphics[width=\textwidth]{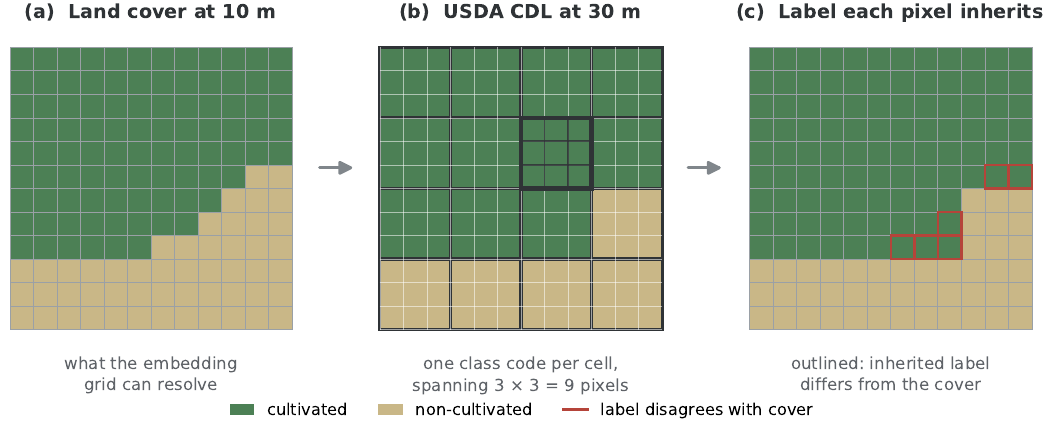}
\caption{How a $30\,\mathrm{m}$ CDL label is carried onto the $10\,\mathrm{m}$ embedding grid.
(a) The land cover an embedding pixel can resolve. (b) The CDL assigns a single class code to each
$30\,\mathrm{m}$ cell, outlined in bold for one cell, which covers a $3\times3$ block of embedding
pixels. (c) Nearest-neighbour alignment gives every pixel in that block the same label, so the
label boundary is forced onto the coarse grid; the outlined pixels are those whose inherited label
therefore differs from the cover beneath them. The disagreement is confined to cells that straddle a
field edge. The figure is a schematic drawn to illustrate the mechanism, not measured data.}
\label{fig:gridmismatch}
\end{figure*}

One consequence of this design deserves to be stated explicitly. The CDL is a $30\,\mathrm{m}$
categorical product, whereas the AlphaEarth embeddings are provided on a $10\,\mathrm{m}$ grid. Each
annual CDL raster is therefore aligned to the EPSG:32619 AlphaEarth grid by nearest-neighbour
resampling, which is the Earth Engine default applied when the sampling toolkit reprojects the layer
without requesting any other kernel, with the target origin, extent and pixel centres fixed by the
embedding raster. No interpolation of class codes is permitted. Consequently, three-by-three groups
of $10\,\mathrm{m}$ model pixels inherit the same $30\,\mathrm{m}$ source label
(Figure~\ref{fig:gridmismatch}), and pixel-level
accuracy against the CDL should not be interpreted as independent $10\,\mathrm{m}$ reference
information, nor as direct evidence of boundary accuracy at $10\,\mathrm{m}$. The disagreement this
introduces is not spread evenly: it is concentrated where a $30\,\mathrm{m}$ cell straddles a field
edge, which is precisely where the classifier's own residual errors also sit
(Section~\ref{sec:maps}). The independent
photo-interpretation of Section~\ref{sec:humanval} provides a separate, though spatially limited,
check on this reference mismatch. The cultivated-code list given above, that is CDL codes 1 to 60,
66 to 80 and 196 to 255, is exactly the list implemented in the executable pipeline as its
\texttt{remap\_crops} rule, and the same list is reported in the repository metadata.

\subsection{Patch generation and sampling}
\label{sec:sampling}
The dataset was prepared jointly for two parallel studies of the same area, one using the AlphaEarth
embeddings (reported here) and one using a fine-tuned segmentation model, so that the two remain
directly comparable. The patches were produced with an open Google Earth Engine toolkit
\citep{montefoschi2026gfm}, and the sampling design explains several properties of the dataset.

A regular grid of cells about $8\,\mathrm{km}$ apart is laid over the area of interest, spaced so
that the extracted patches never touch. Keeping the patches apart matters for two reasons: the same
pixel is never sampled twice, and neighbouring patches cannot leak information between the training
and the test sets, which is a common and easily overlooked cause of over-optimistic accuracy. Within
each cell one location is selected by stratified sampling on the binary CDL label, and around that
location a patch of $224\times224$ pixels at $10\,\mathrm{m}$ resolution, about $2.24\,\mathrm{km}$
on a side, is cut out and reprojected onto a common grid (EPSG:32619, the Sentinel-2
$10\,\mathrm{m}$ grid in UTM zone 19N).

\begin{figure}[tb]
\centering
\includegraphics[width=\linewidth]{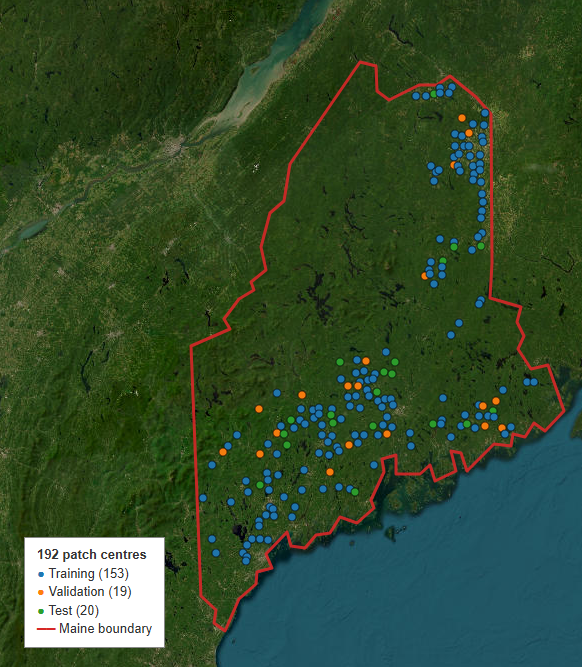}
\caption{Centres of the 192 sampled patches across Maine, coloured by split (training, validation,
test), on an Esri World Imagery basemap; the red line is the state boundary. The patches follow the
agricultural parts of the state, the Aroostook County farmland in the north and the central and
southern valleys, and the three splits are interspersed across the same regions while the
non-contiguous grid keeps every patch several kilometres from its neighbours.}
\label{fig:patch_centers}
\end{figure}

The decisive step for class balance is the acceptance rule: a patch is kept only if at least 10\% of
its pixels are cultivated, and patches below this threshold are discarded. Since cultivated land is
a small minority across Maine, an unconstrained sample would be overwhelmingly non-cultivated; this
rule concentrates the patches on the agricultural parts of the state and is what raises the
cultivated share of the dataset to about 22\%, a balance on which a classifier can be meaningfully
trained and evaluated. Figure~\ref{fig:patch_centers} shows where the 192 accepted patches fall.

This acceptance rule was designed to enrich the evaluation dataset with agricultural areas and
therefore does not produce a probability-based sample representative of Maine as a whole. The 192
sampled patches are appropriate for assessing the model's ability to discriminate between cultivated
and non-cultivated land; their observed class prevalence and overall accuracy should not, however,
be interpreted as statewide estimates of cultivated area or of map accuracy. The separate contiguous
block introduced in Section~\ref{sec:dataset} provides a complementary assessment under spatially
continuous conditions. Nevertheless, that block is also not a random statewide sample, as it was
selected to maximise the achievable cultivated proportion while satisfying the predefined spatial
exclusion constraints.

\subsection{The AlphaEarth embeddings}
\label{sec:aef}
The input to every experiment is the AlphaEarth embedding. For each patch and each year we obtain a
single image of 64 bands from the \texttt{GOOGLE/SATELLITE\_EMBEDDING/V1/ANNUAL} collection in
Google Earth Engine. Every pixel of that image holds a 64-dimensional vector, the model's annual
summary of the $10\,\mathrm{m}$ ground cell beneath it, and it is this vector, rather than any raw
reflectance, that becomes the feature on which the pixel is classified.

A few properties matter for the way we use the embeddings. The product is global and is published as
one layer per year from 2017 onward, which is what makes the year-to-year analysis of
Section~\ref{sec:crossyear} possible; we use the six years from 2018 to 2023. The values are small
numbers of either sign, in our tiles roughly between $-0.51$ and $0.48$, and each 64-vector is
normalised to unit length, so that it lies on the surface of a 63-dimensional unit hypersphere
$\Sph$. This geometric property governs how the embeddings are distributed and is taken up in
Section~\ref{sec:embspace}. In its compact published form each embedding is quantised to one byte
per dimension, that is 64 bytes per pixel, whereas the GeoTIFF tiles we download and work with store
the 64 values in double-precision floating point. Table~\ref{tab:dataset} collects these properties
together with the composition of the dataset.

\begin{figure*}[tb]
\centering
\includegraphics[width=0.92\textwidth]{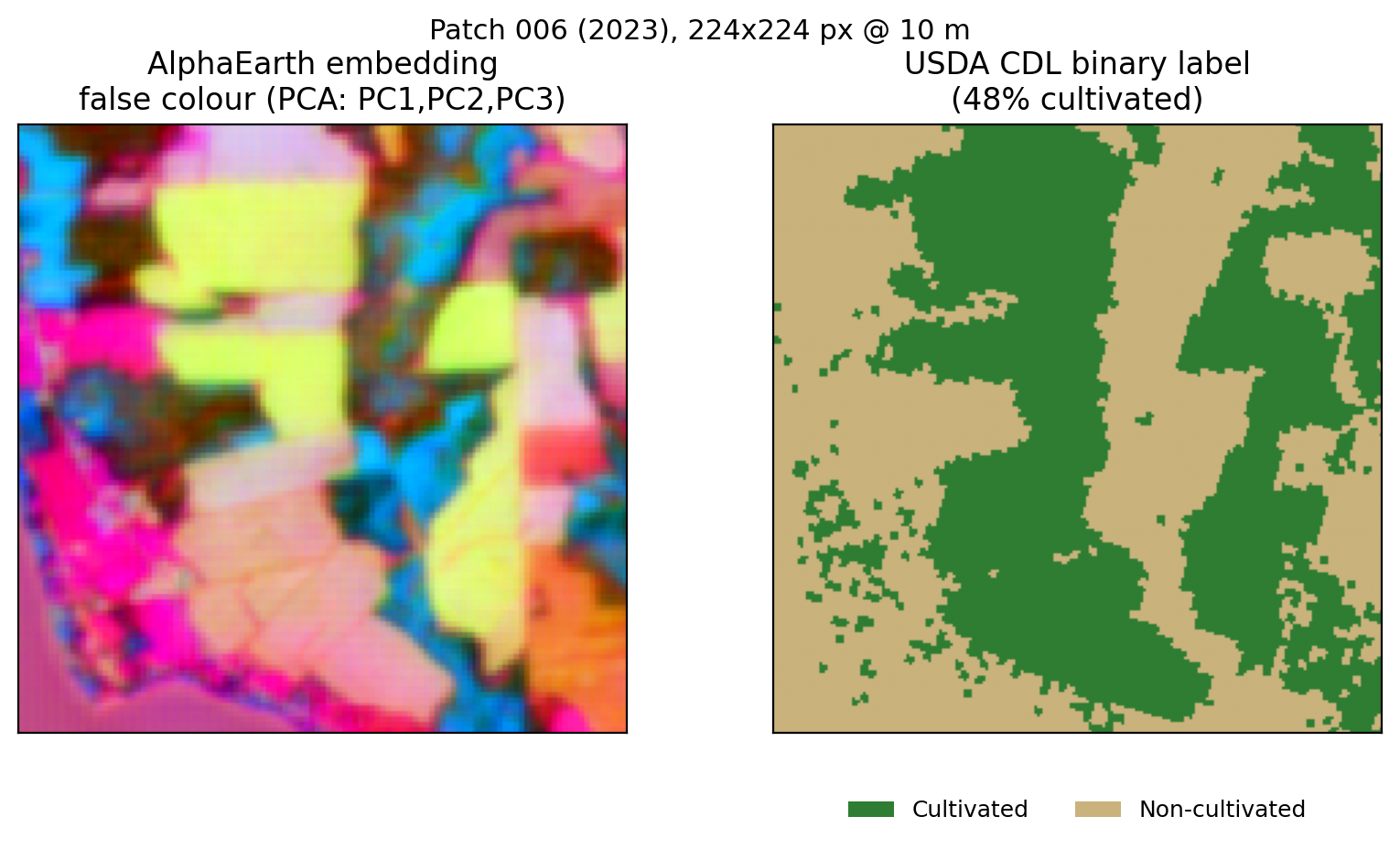}
\caption{One sample patch of the 2023 dataset. Left: the AlphaEarth embedding shown in false colour,
with the first three principal components of the 64 bands mapped to red, green and blue (together
about 57\% of the embedding variance); each channel combines all 64 bands, so the colours carry no
physical meaning. Right: the binary CDL label for the same patch (green: cultivated; tan:
non-cultivated). The patch is $224\times224$ pixels at $10\,\mathrm{m}$ and is 48\% cultivated.}
\label{fig:aef_anchor}
\end{figure*}

Because a 64-band image cannot be shown directly, we visualise an embedding as a false-colour image
built from three channels. Rather than pick three of the sixty-four bands arbitrarily, we summarise
all 64 with a principal component analysis and map the first three principal components to the red,
green and blue channels. The first three components together account for about 57 per cent of the
variance across the embedding, far more than any three individual bands could capture (three
arbitrary bands hold only about 5 per cent, and even the three most variable bands only about 11 per
cent). Each channel is therefore a combination of all 64 numbers, so the colours carry no physical
meaning, but they are the most informative three-channel view available. Figure~\ref{fig:aef_anchor}
does this for one patch and places it next to the binary CDL label of the same patch. Even
compressed to three channels, the field boundaries are clearly visible in the embedding and they
correspond closely to the cultivated areas in the label, which is a first indication that the
embedding already separates fields from their surroundings.

\subsection{The dataset}
\label{sec:dataset}
Applying this sampling over the years 2018 to 2023 yields the dataset used throughout: 192 patches,
each available for all six years, with a 64-band AlphaEarth embedding and the binary CDL label for
every patch and year. The patches are split once into 153 for training, 19 for validation and 20 for
testing, and the same split is used for every year. The training and validation patches are pooled
when fitting the classifiers, while the 20 test patches are held out and used only for the final
evaluation. Table~\ref{tab:dataset} summarises the product specification and the composition for the
reference year 2023. The cultivated share sits between about 18 and 23 per cent across the three
splits, and the overall fraction of about 22\% is the direct effect of the acceptance rule of
Section~\ref{sec:sampling}. Because the sampling grid keeps the patches several kilometres apart,
the training, validation and test patches are spatially separated by construction, which avoids the
spatial leakage that would otherwise inflate the test accuracy. The AlphaEarth embeddings are annual
products and are complete over every patch, so the inputs contain no missing pixels.

\begin{table*}[tb]
\centering
\caption{The AlphaEarth annual embedding product as used here (top) and the composition of the
192-patch dataset for the reference year 2023 (bottom). Pixel counts and cultivated shares are
computed from the binary CDL labels.}
\label{tab:dataset}
\small
\begin{tabular}{ll}
\toprule
\multicolumn{2}{l}{\textbf{Embedding product}}\\
\midrule
Earth Engine collection & \texttt{GOOGLE/SATELLITE\_EMBEDDING/V1/ANNUAL} \\
Embedding dimensions    & 64 bands (A00 to A63) \\
Spatial resolution      & $10\,\mathrm{m}$ \\
Temporal sampling       & one annual layer per year; years used 2018 to 2023 \\
Patch size              & $224\times224$ pixels ($\approx 2.24\,\mathrm{km}$ square) \\
Coordinate system       & EPSG:32619 (UTM zone 19N) \\
Value range (our tiles) & approximately $-0.51$ to $0.48$ \\
Vector norm             & unit length, so each vector lies on $\Sph$ \\
Storage                 & published as int8 (64 bytes per pixel); our tiles float64 \\
\bottomrule
\end{tabular}

\vspace{0.7em}

\begin{tabular}{lrrr}
\toprule
\textbf{Split (2023)} & \textbf{Patches} & \textbf{Pixels} & \textbf{Cultivated (\%)} \\
\midrule
Training    & 153 & 7{,}676{,}928 & 22.35 \\
Validation  & 19  & 953{,}344     & 23.43 \\
Test        & 20  & 1{,}003{,}520 & 18.35 \\
\midrule
Total       & 192 & 9{,}633{,}792 & 22.04 \\
\bottomrule
\end{tabular}
\end{table*}

Alongside the scattered patches we use a separate, continuous test area in southern Maine: a block
of about $11.2\,\mathrm{km}$ by $8.96\,\mathrm{km}$ ($1120\times896$ pixels at $10\,\mathrm{m}$),
tiled into 20 patches in the same format and kept at least $2\,\mathrm{km}$ away from all training
patches, for which we hold the same data for every year from 2018 to 2023. We located it with a
sliding-window search over the state: the region was tiled into 158{,}400 candidate boxes at a
$1\,\mathrm{km}$ stride, every box falling outside Maine or within the $2\,\mathrm{km}$ exclusion
buffer around any training patch was discarded, leaving 56{,}841 valid candidates, and the remaining
boxes were scored by their cultivated fraction. Two targets were searched in parallel, the box
closest to the training-set balance of about 22\% and the box closest to an even split, and both
converged on the same area, at roughly 17\% cultivated. This is a direct consequence of Maine's land
cover: no block of this size anywhere in the valid search region reaches an even split, so the
selected box represents the best achievable class balance rather than a freely chosen target. Its
cultivated fraction is stable across the study period, between about 15\% and 22\% from 2018 to 2023.
This block is the basis of the cross-year check of Section~\ref{sec:crossyear} and of the
photo-interpretation validation of Section~\ref{sec:humanval}.

%% file: sections/03_framework.tex
\section{Embedding-Based Classification Framework}
\label{sec:framework}

\begin{figure*}[tb]
\centering
\includegraphics[width=0.98\textwidth]{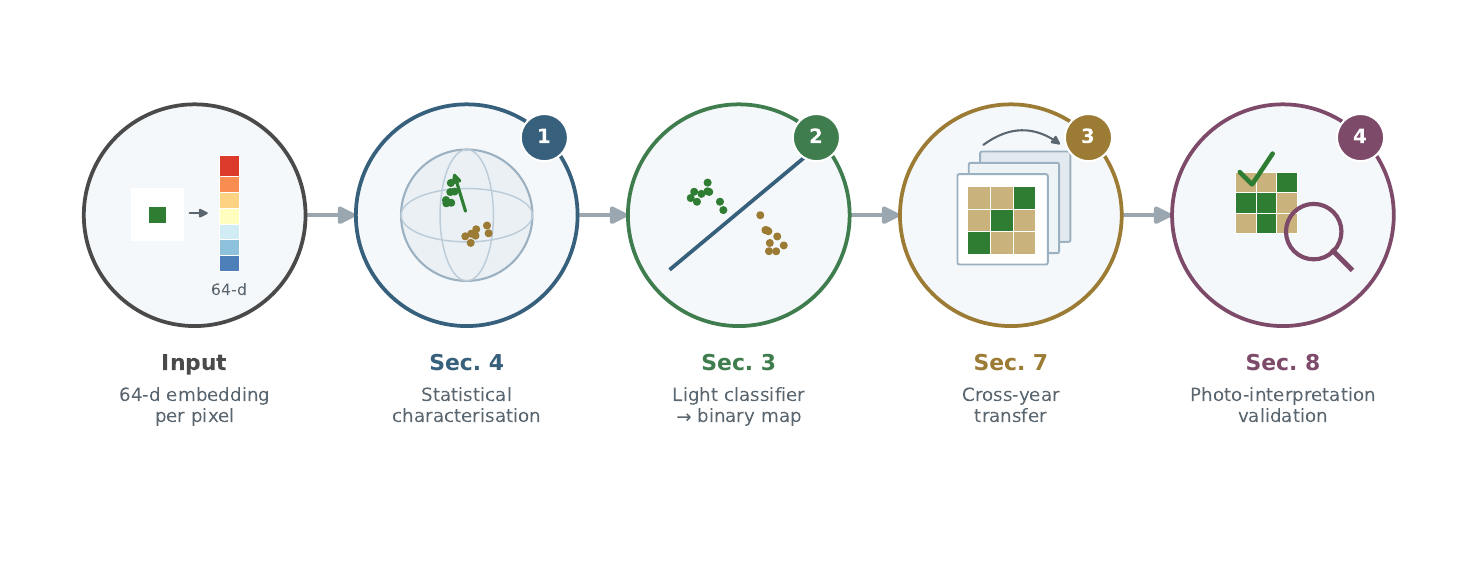}
\caption{Overview of the approach. A 64-dimensional AlphaEarth embedding per pixel is first
characterised statistically (Section~\ref{sec:embspace}), then read by a light classifier into a
binary cultivated map (Section~\ref{sec:framework}); the classifier is then transferred across years
(Section~\ref{sec:crossyear}) and the labels validated by independent photo-interpretation
(Section~\ref{sec:humanval}). The label under each stage gives the section in which that stage is
set out; the corresponding results follow in Sections~\ref{sec:clfresults} to~\ref{sec:humanval}.}
\label{fig:overview}
\end{figure*}

A single decision runs through the whole study. We treat the embedding as the representation and
keep the classifier on top of it deliberately simple. We cannot fine-tune the foundation model
itself, because AlphaEarth is released only as precomputed embeddings and not as an open, trainable
model; and, even setting that aside, we deliberately do not train a deep network of our own. We take
the released embeddings as they are and fit light, standard classifiers to them. This makes the
embedding, rather than the classifier, the object of study, and it raises a question that has to be
answered before any classification: what do these 64 numbers actually look like, and which
statistical tools are even valid on them? The work is therefore organised in two halves, summarised
in Figure~\ref{fig:overview}. Section~\ref{sec:embspace} characterises the embedding space itself;
the present section and those that follow put it to work.

\subsection{Problem set-up}
The classification is binary and per pixel. Each pixel is described by its 64-dimensional embedding
and carries the binary CDL label of Section~\ref{sec:cdlbinary}, and the task is to predict that
label from the embedding. To train and evaluate the classifiers we flatten the patches into a table:
every pixel of every patch becomes one row of 64 columns, and the patches of a split are stacked
into a single matrix $\mathbf{X}\in\mathbb{R}^{n\times 64}$ with a label vector
$\mathbf{y}\in\{1,2\}^{n}$. Invalid pixels, those with a missing embedding or a label that is
neither cultivated nor non-cultivated, are masked out before stacking. The pooled training patches
give about 8.6 million labelled pixels, at the natural class balance of roughly 77.5\%
non-cultivated to 22.5\% cultivated; the 20 test patches, about one million pixels, are held out.

Working per pixel, rather than feeding whole patches to a convolutional network, is a deliberate
choice and it follows directly from what the embedding is. The AlphaEarth embedding was trained to
encode the spatial and temporal context around each pixel, so the 64 numbers at a single location
already summarise its neighbourhood and its year. A classifier placed on top of the embedding
therefore does not need to look at the surrounding pixels itself; reading each pixel's vector
independently is the standard and sufficient recipe, and it keeps the classifier light, which is the
premise of the whole approach. It also means the choice of classifier becomes a question about the
embedding rather than about spatial modelling, which is what makes the comparison of classifiers
below informative. Although the classifiers operate on individual pixels, neighbouring pixels within
a patch are spatially autocorrelated, so the pixel counts should not be interpreted as counts of
fully independent samples; spatial generalisation is assessed instead through held-out patches that
are spatially separated from the training data.

\subsection{Two training regimes: balanced sample against full pool}
\label{sec:regimes}
We train the classifiers under two regimes, and the contrast between them is itself one of the
questions of this paper. The first is a small \emph{balanced sample}: 30{,}000 pixels drawn from
each class, 60{,}000 in total, from the pooled training patches. Drawing equal numbers from the two
classes counteracts the natural imbalance, so the classifier is not simply rewarded for predicting
the majority class, and 60{,}000 pixels is small enough to train even the heavier classifiers, such
as the support vector machine, in seconds. The second regime is the \emph{full pool}: all of the
roughly 8.6 million training pixels, left at their natural balance. Training the same classifiers
both ways tests how much data they actually need. If a model trained on 60{,}000 pixels is almost as
accurate as the same model trained on 8.6 million, then the embedding is informative enough that a
small balanced sample suffices. Throughout, the 20 test patches are never used in either regime.

We use the term \emph{pixel-sample efficiency} for this comparison. The 60{,}000 observations
are labelled pixels inherited from the CDL and sampled from the training patches; they are not
60{,}000 independent manual annotations. The experiment therefore measures how many pixel rows the
downstream classifier needs once a labelled raster is already available, not the cost of building a
new reference dataset from field surveys.

\subsection{Classifiers}
\label{sec:classifiers}
We do not use a single classifier but a graded set, ordered from the simplest possible model to
progressively more flexible ones. The purpose of the ladder is diagnostic: if a one-line linear
model is already almost as good as a large tree ensemble, that tells us the embedding has done most
of the work and the classifier is secondary, which is precisely the claim we set out to examine.
Table~\ref{tab:hyperparams} collects the settings.

Those settings were fixed in advance of any evaluation. They are declared once, as conventional
values for models of this kind, in a version-controlled configuration file of the released pipeline,
and no hyperparameter search of any form was carried out: the codebase contains no grid search,
randomised search or cross-validation procedure. No data therefore informed the choice, and in
particular the 20 test patches were used neither for hyperparameter tuning, nor for
decision-threshold selection, nor for model selection. Because the stacking meta-learner relies on
an internal pixel-level split, its performance is interpreted as a secondary diagnostic rather than
as evidence of spatial generalisation, and the main conclusions rest on the individual classifiers
evaluated on spatially separated test patches. The cost of not searching is that the accuracies
reported below are not optimised ones, and should be read as a floor rather than a ceiling: a tuned
model would plausibly do slightly better. That cost is worth paying here, because the comparison we
care about is between classifiers rather than against a leaderboard, and holding every model at
conventional untuned settings means their near-equal performance cannot be an artefact of unequal
tuning effort.

\emph{Logistic regression} is the linear baseline. It draws a single hyperplane through the
64-dimensional space and reads the probability of cultivated from which side of it a pixel falls,
\begin{equation}
p(\text{cultivated}\mid\mathbf{x}) \;=\; \sigma(\mathbf{w}^{\!\top}\mathbf{x}+b), \qquad
\sigma(z)=\frac{1}{1+e^{-z}},
\label{eq:logreg}
\end{equation}
so there are only 65 parameters to learn. We standardise each dimension first and fit by minimising
the $L_2$-regularised cross-entropy. Logistic regression can only separate the classes with a flat
boundary, so if it performs well that is strong evidence that the embedding is close to linearly
separable for this task, with the foundation model, not the classifier, providing the separation.

\emph{Random forest} \citep{breiman2001random} grows many decision trees, each on a bootstrap sample
of the pixels and with only a random subset of the 64 dimensions considered at each split, and
averages their votes. Because the trees are grown on different samples and feature subsets they are
largely independent, and averaging them reduces variance. A random forest captures non-linear
interactions between dimensions without being told about them and needs no feature scaling, but its
trees are fitted independently and so the forest never corrects its own errors, which is the gap
that boosting fills.

\emph{Gradient boosting} also uses trees, but grows them in sequence, each new tree fitted to the
errors the previous trees still make, so that the score after $T$ rounds is an additive sum of small
trees scaled by a learning rate $\eta$. We use two implementations that differ mainly in how each
tree is grown: LightGBM \citep{ke2017lightgbm} grows leaf-wise, repeatedly splitting whichever leaf
promises the largest loss reduction, and XGBoost \citep{chen2016xgboost} grows level-wise to a fixed
depth and folds the regularisation directly into the split gain it maximises. On our data the two
are effectively tied. For completeness we also include a support vector machine with a radial basis
kernel and a $k$-nearest-neighbours classifier, but only in the balanced-sample regime, since kernel
SVM training grows between quadratically and cubically with the number of pixels and
$k$-nearest neighbours must store and search the entire training set at prediction time.

Because the individual classifiers may make different mistakes, we also combine them, in both cases
on the same balanced 60{,}000-pixel sample as the sampled single classifiers. \emph{Soft voting}
averages the cultivated probabilities of the four base models and thresholds at $0.5$.
\emph{Stacking} instead learns how much to trust each model: the balanced sample is divided by a
stratified random split at the pixel level into a development part (80\%, used to fit the four base
classifiers) and a held-out part (20\%, on which the bases are scored and a logistic-regression
meta-learner is fitted to their predicted probabilities). Scoring the bases on pixels they have
never seen gives the meta-learner realistic, out-of-sample predictions to weigh. This split is at
the pixel level rather than the patch level, so neighbouring pixels of the same patch can fall on
both sides of it; a patch-grouped split would be the cleaner choice for the combiner. It governs
only the internal training of the combiner, since all classifiers are ultimately compared on the 20
held-out test patches, which are spatially separate from the training pool and share no pixels with
it.

\begin{table*}[tb]
\centering
\caption{Classifier settings in the two training regimes. A dash means the model is not used in that
regime. All models use random seed 42. Logistic regression, the random forest, the support vector
machine and $k$-nearest neighbours are implemented in scikit-learn \citep{pedregosa2011scikit};
LightGBM and XGBoost use their own libraries.}
\label{tab:hyperparams}
\small
\setlength{\tabcolsep}{5pt}
\begin{tabular}{lll}
\toprule
\textbf{Classifier} & \textbf{Full pool ($\sim$8.6\,M px)} & \textbf{Balanced sample (60\,k px)} \\
\midrule
Logistic regression & standardise $+$ $L_2$ ($C{=}1$) & standardise $+$ $L_2$ ($C{=}1$) \\
Random forest & 200 trees, depth 22 & 250 trees, depth 20 \\
LightGBM & 400 trees, $\eta{=}0.05$, 31 leaves & 450 trees, $\eta{=}0.03$, 255 leaves \\
XGBoost & 400 trees, $\eta{=}0.05$, depth 6, $\lambda{=}1.0$ & 450 trees, $\eta{=}0.03$, depth 12, $\lambda{=}0.5$ \\
SVM (RBF) & -- & $C{=}2.0$ \\
$k$-nearest neighbours & -- & $k{=}5$ \\
\bottomrule
\end{tabular}
\end{table*}

\subsection{Evaluation protocol and metrics}
\label{sec:metrics}
Every classifier and ensemble is evaluated on the 20 held-out test patches, which are never used in
any fitting and, because of the sampling grid of Section~\ref{sec:sampling}, are spatially separated
from the training patches so that no pixel and no immediate neighbourhood is shared. This is what
makes the test accuracy an estimate of performance on genuinely new ground rather than a measure of
memorisation. To be explicit about how the splits are used: the 153 training and 19 validation
patches are pooled into the fitting pool, so the validation patches enter model fitting rather than
serving as a separate patch-level selection set, and the classifiers are compared on the held-out
test patches. Because the leading classifiers differ by only about a tenth of a percentage point,
the conclusions do not depend on which of them is retained.

We report several metrics because the task is imbalanced: with only about 22\% of pixels cultivated,
a trivial model that calls everything non-cultivated already scores about 78\% overall accuracy.
Alongside overall accuracy we therefore report the per-class precision and recall, the
macro-averaged $F_1$ score, the balanced accuracy
\begin{equation}
\mathrm{BA} = \tfrac{1}{2}\bigl(R_{\text{cult}} + R_{\text{non}}\bigr),
\label{eq:balacc}
\end{equation}
the unweighted mean of the two classes' recalls, and the cultivated intersection-over-union
$\mathrm{IoU}_{\text{cult}} = TP/(TP+FP+FN)$. Because each class counts equally in
Equation~\ref{eq:balacc} regardless of its size, a model that simply predicts the majority class
scores only $0.5$, which makes the balanced accuracy a fairer single number than overall accuracy on
this task. All experiments use the fixed random seed 42 so that the splits, samples and model fits
are reproducible.

The independent sampling unit for spatial generalisation is the held-out patch, not the pixel.
We therefore use the pixel-level metrics as descriptive summaries over the held-out area, and avoid
interpreting the approximately one million test pixels as one million independent observations.
Formal paired inference is used only in Section~\ref{sec:humanval}, where competing maps are
evaluated on the same independently sampled reference points. Differences of a few tenths of a
percentage point between classifiers should consequently be read as practically small, unless they
are stable across patches or supported by an explicit uncertainty analysis.

%% file: sections/04_embedding_space.tex
\section{Embedding-Space Analysis}
\label{sec:embspace}

Before fitting any classifier we study the embeddings themselves. There are two reasons for doing
this first. The practical reason is that the analysis decides which tools are valid further down:
several of the tests and models we might reach for, and the distance an algorithm uses to compare
two pixels, assume a particular shape for the data, and if that shape does not hold we have to
choose differently. The second reason is that a careful look at the embedding space gives a first,
classifier-free picture of how hard the task is, by measuring directly how far apart the cultivated
and non-cultivated pixels sit.

One caveat applies throughout. Because neighbouring pixels within the same patch are spatially
autocorrelated, and because the sample sizes are very large, the p-values reported below are used
mainly as diagnostics, while effect sizes, angular separability and held-out classification
performance are treated as the more informative measures.

\subsection{Directional geometry}
\label{sec:geometry}
Each AlphaEarth pixel is a vector $\mathbf{e}\in\mathbb{R}^{64}$, and the published embeddings are
$L_2$-normalised to unit length, so every embedding lies on the surface of the unit hypersphere
$\Sph=\{\mathbf{x}\in\mathbb{R}^{64}:\lVert\mathbf{x}\rVert_2=1\}$. We confirmed this on our own
tiles: the mean Euclidean length of a sampled embedding is $1.000$ to three decimal places. The
consequence is geometric and important for everything that follows. Because all vectors have the
same length, the magnitude of an embedding carries no information and only its \emph{direction}
does, so the natural way to compare two pixels is the angle between them, measured by the cosine
similarity $s(\mathbf{e}_i,\mathbf{e}_j)=\mathbf{e}_i\cdot\mathbf{e}_j$, where the equality with the
inner product holds precisely because the vectors are unit length.

This also bears on the statistics we are allowed to use. A multivariate Gaussian places its mass
throughout $\mathbb{R}^{64}$, including regions off the sphere where no embedding can ever fall, so
it is geometrically mismatched to data confined to $\Sph$. The same constraint has a stricter,
one-dimensional consequence: the unit norm forces every single coordinate into the bounded interval
$[-1,1]$, because $e_j^2\le\sum_k e_k^2=1$, and in our tiles into a much narrower range still
(Table~\ref{tab:dataset}). A bounded quantity cannot be exactly normal, so the non-normality
documented below is in part a structural consequence of the spherical geometry rather than a
surprise. The matching family is directional statistics, which model points on a sphere directly and
replace ordinary distance with angle. We stress that unit normalisation makes a directional
treatment \emph{appropriate}, but it does not by itself guarantee that any single directional model
fits; that has to be tested empirically.

\subsection{Concentration and the von Mises-Fisher model}
\label{sec:vmf}
The natural first model for data on $\Sph$ is the directional analogue of the Gaussian, the von
Mises-Fisher (vMF) distribution \citep{mardia2000directional}, whose density
$f(\mathbf{x}\mid\boldsymbol{\mu},\kappa)=C_d(\kappa)\exp(\kappa\,\boldsymbol{\mu}^{\!\top}\mathbf{x})$
is governed by a mean direction $\boldsymbol{\mu}$ and a concentration $\kappa$ that plays the role
the inverse variance plays for a Gaussian. We estimate $\kappa$ from the mean resultant length
$\bar{R}=\lVert\frac{1}{n}\sum_i\mathbf{x}_i\rVert$, the length of the \emph{average} of $n$ unit
vectors, which is short when the directions disagree and cancel and long when they agree, using the
closed-form approximation of \citet{banerjee2005clustering} with $d=64$.

It matters \emph{which} set of embeddings is being measured. AlphaEarth is trained with a
batch-uniformity objective whose explicit purpose is to spread the embeddings as evenly as possible
over $\Sph$ \citep{alphaearth2025}, so taken over the whole Earth the distribution is close to
uniform, which by definition means a near-zero concentration. The high concentrations we report are
not in tension with this: we never measure the global set, but the embeddings of one region split
into two land-cover classes, and a subset that is globally a thin slice of an overall-uniform sphere
can perfectly well have a large $\kappa$ of its own. Within each class the concentration is high, of
the order of $10^2$ (Table~\ref{tab:geometry}), corresponding to tight angular clusters. The
non-cultivated class is slightly more concentrated ($\hat\kappa\approx159$) than the cultivated class
($\hat\kappa\approx137$); a plausible reading, which we do not test directly, is that the
non-cultivated class is dominated by spectrally homogeneous forest while the cultivated class mixes
several crop types. The combined estimate ($\hat\kappa\approx125$) is lower than either class alone,
exactly as expected when two slightly offset clusters are pooled. The per-class estimate computed on
a 50{,}000-pixel subsample ($137.1$) is very close to the full-pool value ($136.9$), and the estimate
settles from only a few hundred pixels, so it appears to reflect a stable property of the data rather
than a sampling artefact.

Two cautions apply. AlphaEarth's own use of the vMF is a \emph{training-time} mechanism with an
imposed concentration, a stochastic bottleneck on the whole 64-dimensional unit vector; it does not
assert that the empirical distribution of the released embeddings is itself vMF, and that empirical
question is the one we take up here. Second, a single vMF describes a class as one round blob with
exactly one preferred direction, which is faithful only if the class really is one cluster. The
cultivated class is not one thing, it is potatoes, hay, blueberries and so on, and each crop may
point in its own direction, so a mixture of several vMF components may describe it better. Our
low-dimensional projections do show such sub-clusters, so we use $\kappa$ as a descriptive summary
of concentration rather than as an established model.

\subsection{Normality and marginal shape}
\label{sec:normality}
The analyses available to us split into two families: one that assumes approximately normal data,
including parametric two-sample tests and Gaussian-implied distances, and one that is
distribution-free. Normality testing is therefore used here as a diagnostic step that decides which
family is legitimate. We test each of the 64 dimensions, separately for the two classes, with three
complementary tests, controlling the family-wise error rate over the $64\times2=128$ tests with a
Bonferroni correction: Shapiro-Wilk on a 5{,}000-pixel subsample, D'Agostino-Pearson $K^2$ and
Anderson-Darling on a 50{,}000-pixel sample.

Normality is rejected almost everywhere. Of the 128 dimension-class combinations, 126 reject the
null on all three tests and the remaining two reject on at least one, with Shapiro-Wilk p-values many
orders of magnitude below the threshold and widespread non-zero skewness and excess kurtosis.
Fitting a panel of seven candidate distributions per dimension (normal, Student's $t$, Laplace,
logistic, generalised normal, skew-normal and Cauchy) and selecting by the Akaike information
criterion confirms this quantitatively: the normal is never the preferred model, the skew-normal is
the single most frequent best fit (73 of the 128 cases), followed by the generalised normal (46),
with a handful of Student's $t$ (8) and one logistic fit. Averaged over the cases, the improvement of
the best fit over the normal is large, with a mean $\Delta\mathrm{AIC}$ of about 2{,}400, where a
difference above ten already counts as decisive. The dominant departure is therefore asymmetry, not
heavy tails: the marginals are best described as mildly skewed bell shapes. These fitted
distributions are used as empirical summaries of marginal shape, not as exact generative models,
since the coordinates are bounded by the unit-norm constraint whereas most of the candidates have
unbounded support.

One point is worth stating clearly, because it is easy to misread. Failing the normality test does
not mean a dimension is uninformative. The rejection of exact marginal normality does not imply
that linear classifiers, or large-sample comparisons of class means, are inappropriate. Rather, it
indicates that a multivariate Gaussian model does not adequately describe the distribution of the
unit-normalised embeddings, and that, given the very large sample size, marginal-test p-values alone
provide limited information about practical class separation. We therefore place greater emphasis on
effect sizes, distribution-free comparisons, directional geometry, and predictive performance on
held-out data. Gaussian-based analyses are retained only as secondary large-sample consistency
checks.

\subsection{Per-dimension signal and class separability}
\label{sec:perdim}
We now ask whether all 64 dimensions actually matter for telling cultivated from non-cultivated
land, or whether only a handful do the separating. The qualification matters: a dimension that is
uninformative for this binary distinction is not noise in general, and may well be exactly what
separates two forest types or two crop types in another task.

Looking at the per-dimension mean difference alone is not enough, because a difference in means is
only meaningful relative to the spread of the data. We therefore summarise each dimension by Cohen's
standardised mean difference $d_{\mathrm{C},j}$ \citep{cohen1988statistical}, which expresses the gap
between the two class means in units of the data's own spread and so does not change with sample
size. This is preferable to a p-value here: at our sample sizes almost any non-zero difference is
statistically significant, so a p-value would flag essentially every dimension and tell us nothing
about which ones actually help.

\begin{figure*}[tb]
\centering
\includegraphics[width=0.94\textwidth]{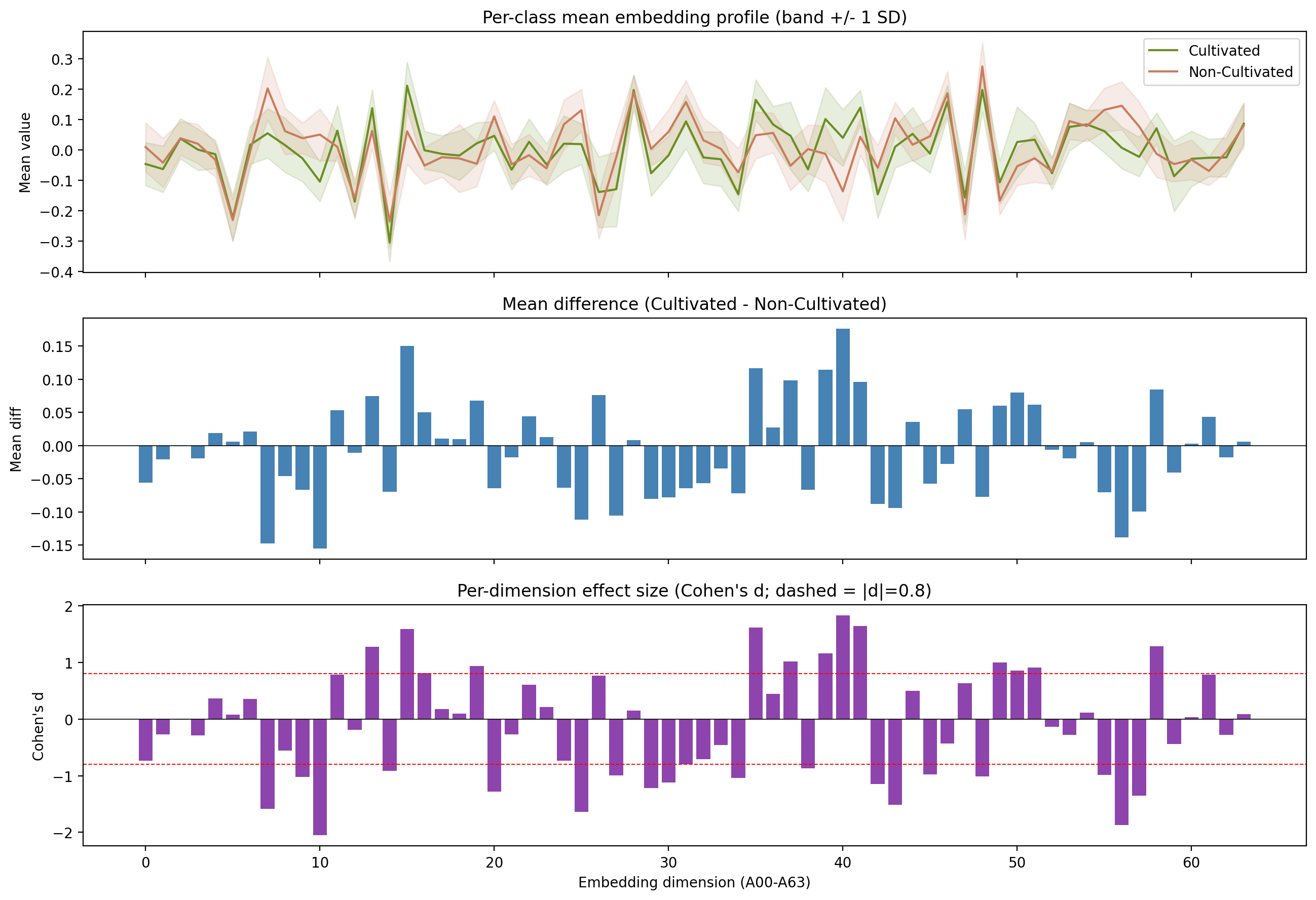}
\caption{Per-dimension profiles for the two classes. Top: class mean profiles with $\pm1$
standard-deviation bands; the bands overlap heavily on every dimension. Middle: per-dimension mean
difference (cultivated minus non-cultivated), small but systematic. Bottom: per-dimension effect
size, Cohen's $d_{\mathrm{C},j}$, with the large-effect line $|d|=0.8$ dashed. Many dimensions reach
a large effect, but none is individually decisive.}
\label{fig:mean_profiles}
\end{figure*}

The separating signal is spread across the embedding rather than held in a few coordinates
(Figure~\ref{fig:mean_profiles}). Across the 64 dimensions, 31 reach a large effect
($|d_{\mathrm{C}}|\ge0.8$), 10 a medium and 13 a small one, and only 10 are effectively negligible,
with a mean absolute effect of $0.80$ and a maximum near $2.0$. Yet the two class mean profiles track
each other closely and their $\pm1$ standard-deviation bands overlap heavily on every dimension, so
even the strongest single dimension leaves the classes far from separable on its own. A large effect
size is not the same as a dimension that decides the class: at $d_{\mathrm{C}}=0.8$ the two class
means are separated by less than one standard deviation, so a threshold rule on that one dimension
would still misclassify a large fraction of pixels. The two statements are consistent, and together
they say that the distinction is encoded \emph{collectively}, across many weakly-to-moderately
informative dimensions.

Distribution-free tests agree that almost every dimension carries class signal. Using the
Mann-Whitney $U$, Kolmogorov-Smirnov and Brunner-Munzel tests for location and distribution, and a
median-based Levene test for spread, all at the Bonferroni-corrected level, 61 of the 64 dimensions
differ between the classes in both location and spread, two differ in location only, one in spread
only, and none fails to differ on both. Welch's $t$-test, run as a parametric cross-check and valid
here by the central limit theorem despite the non-normality of the data, agrees on essentially every
dimension. As above, the practical message is read from the effect sizes rather than from
significance.

\subsection{Angular geometry and a low-capacity supervised baseline}
\label{sec:cosine}
The tests so far treat the 64 dimensions one at a time. The cosine analysis instead measures
separability in the way the geometry of Section~\ref{sec:geometry} suggests, by angle between whole
vectors, and it needs no assumption about the marginal distributions. For each class we form the
centroid, the normalised mean direction of that class's embeddings, and summarise the geometry by
the within-class cohesion (the mean cosine of a class's pixels to their own centroid), the
cross-class similarity (to the other class's centroid), their difference, and the angle between the
two centroids. We compute these with both the mean and the median centroid and check that the two
agree.

The picture is of two classes that sit close together on the sphere but remain distinguishable
(Table~\ref{tab:geometry}). The within-class cohesion is $0.808$, clearly above the cross-class
similarity of $0.599$, giving a separation gap of $0.209$, and the two centroids are separated by an
angle of $42.2^{\circ}$. All these similarities are positive, so both classes occupy the same broad
region of the sphere rather than pointing in opposite directions, which is why the task is
non-trivial. 
ev{Even so, a nearest-class-centroid rule, which estimates one labelled centroid per
class but performs no iterative parameter fitting, classifies $90.2\%$ of pixels correctly. This
should be read as a low-capacity \emph{supervised} baseline rather than a label-free method. Its
performance shows that a large fraction of the class separation is already expressed by the
embedding geometry, while the remaining gain from trained classifiers reflects a more flexible
decision boundary, and it fixes a baseline the trained classifiers must clearly beat.}

\begin{table}[tb]
\centering
\caption{Directional geometry of the two classes on $\Sph$, computed on the 192-patch dataset.
Higher $\hat\kappa$ means a tighter angular cluster. Cohesion is the mean cosine of a class's pixels
to their own centroid and cross-class similarity the mean cosine to the other class's centroid; the
nearest-class-centroid rule estimates one labelled centroid per class and fits no further parameters.}
\label{tab:geometry}
\small
\setlength{\tabcolsep}{4pt}
\begin{tabular}{lr}
\toprule
\textbf{Quantity} & \textbf{Value} \\
\midrule
$\hat\kappa$, cultivated & 136.9 \\
$\hat\kappa$, non-cultivated & 158.5 \\
$\hat\kappa$, combined & 125.1 \\
\midrule
Within-class cohesion & 0.808 \\
Cross-class similarity & 0.599 \\
Separation gap & 0.209 \\
Centroid angle & $42.2^{\circ}$ \\

ev{Nearest-class-centroid accuracy} & 90.2\% \\
\bottomrule
\end{tabular}
\end{table}

\subsection{Can the 64 dimensions be compressed?}
\label{sec:dimred}
Finally we test directly whether the embedding can be reduced without losing accuracy, from an
unsupervised direction (principal component analysis, which compresses by variance) and a supervised
one (feature selection, which keeps only the dimensions that separate the classes).

One point on the set-up, because it affects how the numbers below should be read. The experiments in
this subsection, and the tangent-space comparison that closes it, are run on the class-balanced
sample of 100{,}000 pixels (50{,}000 per class) used throughout the embedding-space analysis, with a
70/30 split inside that sample and lighter classifier settings than those of
Table~\ref{tab:hyperparams}. They therefore use neither of the training regimes of
Section~\ref{sec:regimes} and are not evaluated on the held-out test patches. Their absolute
accuracies are consequently a little lower than those of Table~\ref{tab:classifiers} and should not
be compared with it; what is meaningful here is the change in accuracy as dimensions are removed,
which is measured under identical conditions throughout.

\begin{figure*}[tb]
\centering
\includegraphics[width=0.94\textwidth]{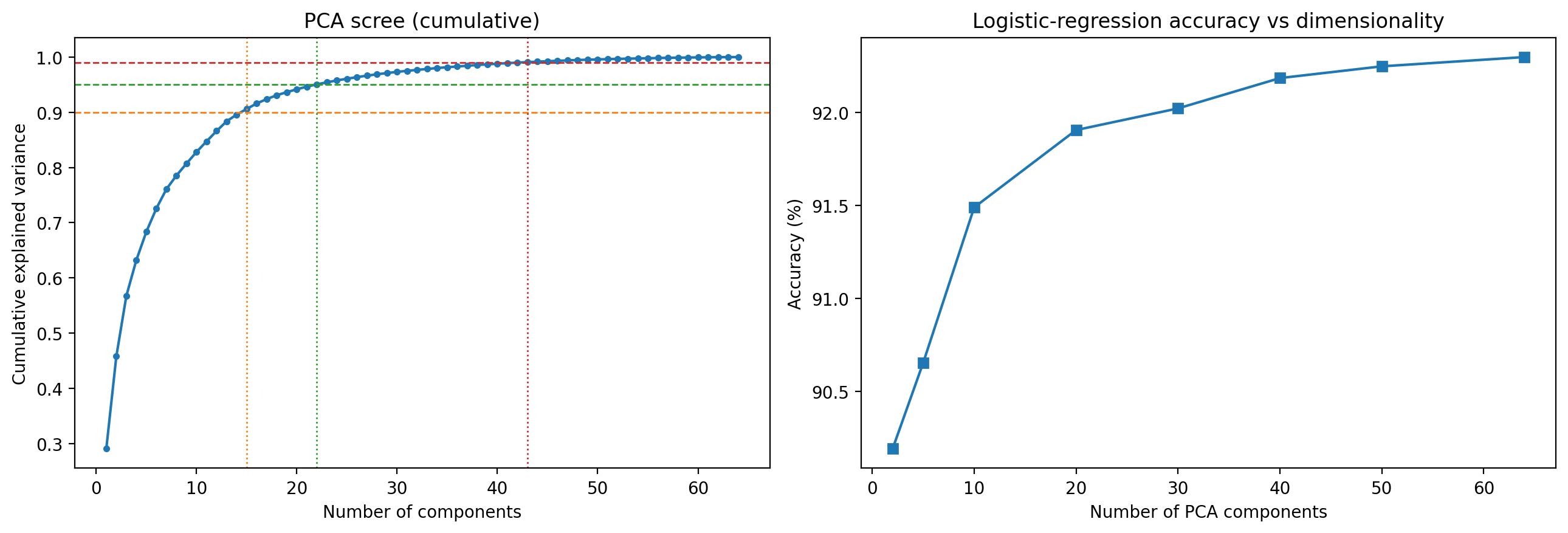}
\caption{Left: cumulative variance explained by the principal components, with the 90\%, 95\% and
99\% levels marked; the variance saturates after roughly 22 components. Right: logistic-regression
accuracy as a function of the number of principal components used; accuracy keeps rising past the
point where the variance has saturated, because low-variance directions still carry class signal.}
\label{fig:pca}
\end{figure*}

The variance concentrates quickly: about 15 principal components capture 90\% of the total variance,
22 capture 95\% and 43 capture 99\% (Figure~\ref{fig:pca}, left). Variance is not the same as class
information, however. A logistic regression trained on the first $m$ principal components improves
monotonically as $m$ grows, from $90.2\%$ on the first two components to $92.3\%$ on all 64
(Figure~\ref{fig:pca}, right), and is still creeping upward at the end: the last components, which
carry almost no variance, still add accuracy, because principal component analysis orders directions
by variance and not by class separation. The supervised view agrees. Keeping only the 61 separating
dimensions and discarding the other three lowers accuracy slightly rather than leaving it unchanged,
from $92.96\%$ to $92.89\%$ for the random forest and from $92.28\%$ to $92.13\%$ for logistic
regression. There is therefore no subset of the 64 dimensions that can be dropped for free.

This conclusion needs stating carefully, because two recent studies of the embedding space
reach what may look like the opposite result. \citet{benavides2026alphaearth} report that land-cover
classification at 98\% of baseline performance can be reached with as few as 2 to 12 of the 64
dimensions, depending on the class, eight of them for their cropland class, and
\citet{rahman2026geometry} estimate an effective
dimensionality of about 13 and a local intrinsic dimensionality near 10. The difference is the
acceptance criterion rather than the measurement. Those studies ask how few dimensions suffice while
tolerating a small relative loss, and answer that very few do; we ask whether any dimension can be
removed at \emph{no} measurable cost, and find that none can. Both readings describe the same
embedding: the class signal is heavily redundant, yet thinly spread, so that discarding coordinates
is cheap but never free. Read through its correlation structure rather than its task performance,
the space is in fact only moderately redundant: \citet{rahman2026geometry} report that just 47 of
the 2{,}016 dimension pairs exceed a correlation of $0.5$ in absolute value, so most dimensions
carry variance that is not duplicated elsewhere. Within this balanced Maine binary-classification experiment, neither
principal component analysis nor the tested supervised feature-selection procedure improves
accuracy, and retaining all 64 dimensions is the safest choice for the analyses that follow. This is
a task- and protocol-specific result, not a claim that AlphaEarth dimensions are universally
irreducible; where a small accuracy loss is acceptable, the evidence above suggests a much smaller
subset would serve.

There is also a geometric reason to be wary of reduction. The directional analysis rests on the
embeddings lying on $\Sph$; projecting onto a handful of principal components, or dropping
coordinates outright, moves the points off that sphere and breaks the unit-norm property, so the
angular geometry and any directional model no longer strictly apply to the reduced vectors. We
therefore keep all 64 dimensions and use principal component analysis only as a diagnostic and for
the false-colour visualisation of Section~\ref{sec:aef}. For the same reason we tested whether
respecting the curvature of the sphere helps, by projecting the embeddings into the tangent space at
their Fr\'echet mean before classifying; across four classifiers this changed accuracy by at most
$0.03$ percentage points in either direction, well inside run-to-run noise, which is what the high
concentration of Section~\ref{sec:vmf} would predict, since a class that occupies such a small patch
of the sphere is almost flat already. This last result should be scoped carefully.
\citet{rahman2026geometry}, characterising the AlphaEarth manifold over 12.1 million samples across
the conterminous United States, report that tangent spaces rotate substantially from place to place,
with 84\% of sampled locations showing tangent-space angles above $60^{\circ}$ and local-global
principal-component alignment close to the random baseline. Our finding does not contradict theirs:
a single tangent space taken at the Fr\'echet mean of two tightly concentrated classes within one
region is a very different object from the manifold as a whole, and the same subset-versus-global
distinction applies here as it does to $\kappa$ in Section~\ref{sec:vmf}. A single global tangent
space is adequate for this regional task; it would not be for the manifold at large.

Taken together, these results show that the AlphaEarth embedding space is not well described by
simple Gaussian coordinate-wise assumptions, but that it has a clear and stable angular structure.
The cultivated and non-cultivated classes are not separated by one or two dominant dimensions;
instead, the signal is distributed across the full 64-dimensional vector. This justifies carrying
all dimensions into the classification experiments, and it explains why even simple classifiers can
perform competitively.

%% file: sections/05_classification_results.tex
\section{Classification Results}
\label{sec:clfresults}

\subsection{The classifier ladder}
\label{sec:ladder}
The central result of the classification experiments is how little the choice of classifier matters
once the embedding is fixed. Table~\ref{tab:classifiers} reports overall accuracy, macro-averaged
$F_1$ and balanced accuracy on the held-out test patches, for the full-pool models and for the
balanced-sample models. On the full pool the four models span a band of just $0.27$ percentage
points, from $93.48\%$ for plain logistic regression to $93.75\%$ for the random forest and XGBoost,
with macro-$F_1$ around $0.89$ throughout. A single linear hyperplane through the 64-dimensional
space is therefore almost as accurate as a large gradient-boosted ensemble, which indicates that the
embedding space is close to linearly separable for this task and that the foundation model, rather
than the classifier, provides the separation. This connects back to how the embeddings are trained:
the batch-uniformity objective spreads the representation over the hypersphere and prevents
collapse, so the discriminative structure is largely present in the embedding geometry already and
little is left for the classifier to add. The same point is made from below by the nearest-centroid
baseline of Section~\ref{sec:cosine}: at $90.2\%$ from labelled class centroids alone, with no
iterative parameter fitting, it is already within roughly three points of the trained models.

\begin{table*}[tb]
\centering
\caption{Classification accuracy on the 20 held-out test patches. The full-pool models are trained
on all ${\sim}8.6$ million pixels of the training pool (train plus validation); the balanced-sample
models on 60{,}000 pixels (30{,}000 per class). Balanced accuracy is the mean of the two class
recalls. A dash means the model is not run in that regime.}
\label{tab:classifiers}
\small
\setlength{\tabcolsep}{4.5pt}
\begin{tabular}{@{}lccc@{\hskip 12pt}ccc@{}}
\toprule
 & \multicolumn{3}{c}{\textbf{Full pool}} & \multicolumn{3}{c}{\textbf{Balanced sample}} \\
\cmidrule(r){2-4}\cmidrule(l){5-7}
\textbf{Classifier} & OA & macro-$F_1$ & Bal.\ acc. & OA & macro-$F_1$ & Bal.\ acc. \\
\midrule
Logistic regression & 93.48\% & 0.892 & 89.6\% & 92.05\% & 0.880 & 92.4\% \\
\textbf{Random forest} & \textbf{93.75\%} & \textbf{0.898} & \textbf{90.8\%} & \textbf{92.45\%} & \textbf{0.886} & 92.6\% \\
LightGBM & 93.72\% & 0.896 & 90.0\% & -- & -- & -- \\
XGBoost & 93.75\% & 0.897 & 90.0\% & -- & -- & -- \\
SVM (RBF) & -- & -- & -- & 92.22\% & 0.883 & \textbf{93.0\%} \\
Stacked ensemble & -- & -- & -- & 92.64\% & 0.888 & 92.7\% \\
Nearest class centroid & \multicolumn{3}{c}{90.2\% (OA)} & \multicolumn{3}{c}{--} \\
\bottomrule
\end{tabular}
\end{table*}

The overall accuracies of the classifiers differ by only a few tenths of a percentage point,
whereas performance varies across held-out patches and errors are spatially clustered. We therefore
treat the principal empirical result as equivalence in practical performance among several
lightweight readouts, rather than as evidence that one specific classifier is universally best. The
random forest is selected for the mapping that follows because it combines strong performance,
non-linear capacity, and straightforward implementation; it needs no feature scaling, and it is the
model used unchanged for the cross-year transfer of Section~\ref{sec:crossyear} and for the human
validation of Section~\ref{sec:humanval}.

\subsection{Class imbalance and per-class behaviour}
\label{sec:balanced}
Because only about $18\%$ of the test pixels are cultivated, overall accuracy alone flatters the
result: a trivial model that labelled everything non-cultivated would already score about $82\%$.
The balanced accuracy corrects for this. For the chosen random forest it is $90.8\%$ against
$93.75\%$ overall: the model recovers $95.4\%$ of the non-cultivated pixels and $86.2\%$ of the
cultivated ones. The gap between the two recalls is the expected signature of the imbalance and of
the cultivated class being the harder, more heterogeneous one, but a balanced accuracy close to
$91\%$ shows that the minority class is genuinely being recovered and not sacrificed to the
majority.

The two training regimes trade these quantities off against each other. Under full-pool training
balanced accuracy sits between $89.6\%$ and $90.8\%$, mirroring the overall-accuracy ordering.
Training on the class-balanced 60{,}000-pixel sample instead raises balanced accuracy to between
$92.4\%$ and $93.0\%$, even though it lowers overall accuracy by about one percentage point:
balancing the training set recovers more of the minority cultivated class, whose recall rises from
about $86\%$ to about $93\%$, at the cost of a few more non-cultivated false positives. Which regime
is preferable therefore depends on whether overall agreement or an even-handed treatment of the two
classes matters more for the intended application.

Table~\ref{tab:perclass} gives the full per-class breakdown for the chosen model. The cultivated
class is consistently the harder of the two, with a lower precision ($80.9\%$) and recall ($86.2\%$)
than the non-cultivated class (both above $95\%$), so its $F_1$ ($0.835$) trails the non-cultivated
$F_1$ ($0.961$). The confusion matrix behind these numbers has $781{,}982$ non-cultivated and
$158{,}796$ cultivated pixels correctly classified, against $37{,}375$ cultivated false positives and
$25{,}367$ cultivated false negatives, so the residual error is split between the two kinds of
cultivated mistake rather than dominated by either.

\begin{table*}[tb]
\centering
\caption{Per-class performance of the chosen random forest on the held-out 2023 test patches
(full-pool training). Balanced accuracy equals the macro-averaged recall.}
\label{tab:perclass}
\small
\begin{tabular}{lrrrr}
\toprule
\textbf{Class} & \textbf{Precision} & \textbf{Recall} & \textbf{$F_1$} & \textbf{Pixels} \\
\midrule
Non-cultivated & 96.9\% & 95.4\% & 0.961 & 819{,}357 \\
Cultivated & 80.9\% & 86.2\% & 0.835 & 184{,}163 \\
\midrule
Macro average & 88.9\% & 90.8\% & 0.898 & 1{,}003{,}520 \\
\bottomrule
\multicolumn{5}{l}{\small Overall accuracy $93.75\%$; balanced accuracy $90.8\%$.} \\
\end{tabular}
\end{table*}

\subsection{Predictions in the map domain}
\label{sec:maps}
\begin{figure*}[tb]
\centering
\includegraphics[width=\textwidth]{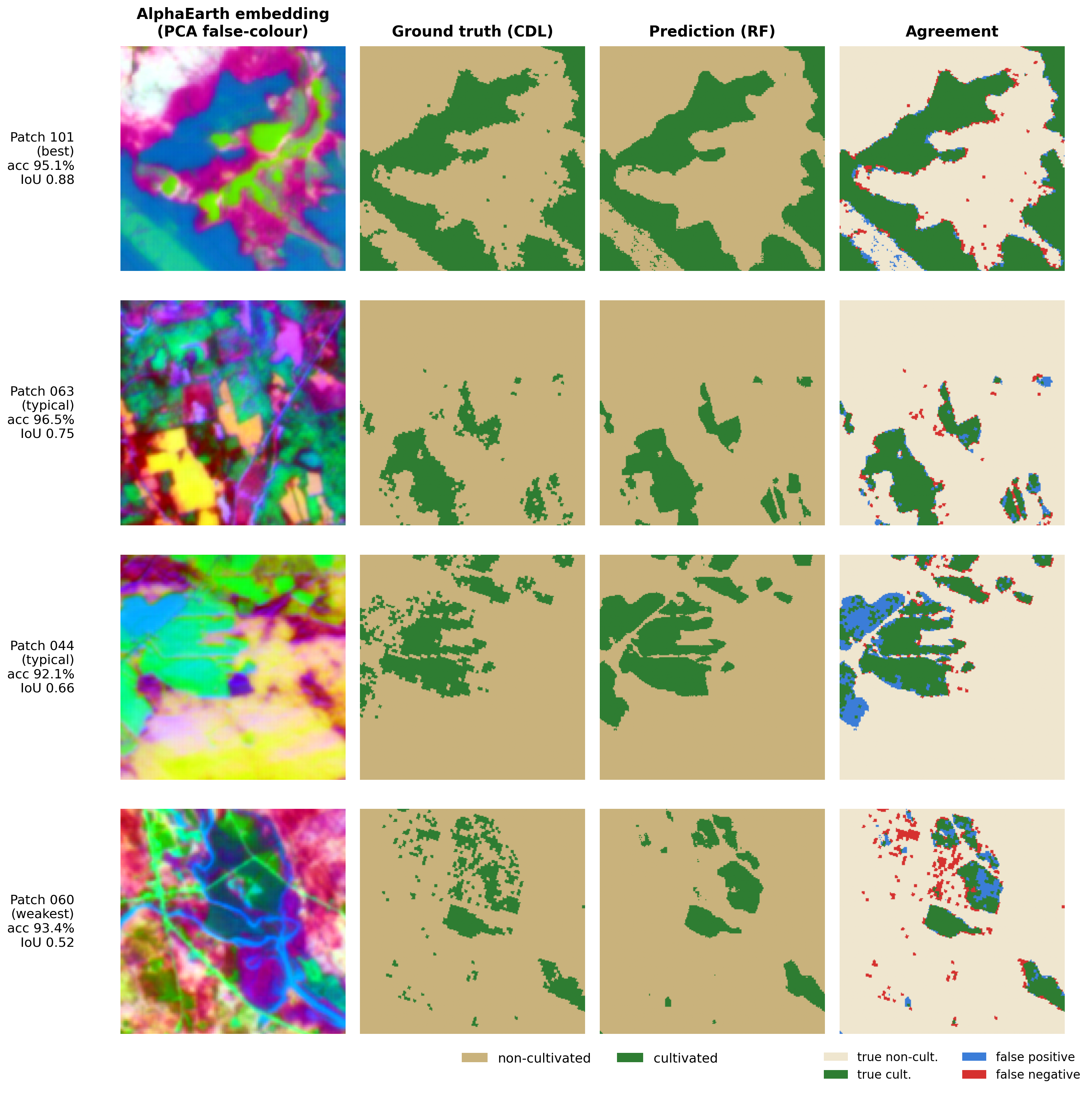}
\caption{Random-forest predictions on four held-out 2023 test patches, ordered from the strongest to
the weakest by cultivated intersection-over-union. Columns, left to right: the AlphaEarth embedding
as a principal-component false-colour; the CDL ground truth; the random-forest prediction; and the
agreement map (true non-cultivated, true cultivated, false positive, false negative). The prediction
closely tracks the ground truth, with errors concentrated on field boundaries and small parcels.}
\label{fig:pred_patches}
\end{figure*}

The aggregate numbers are easiest to interpret when the predictions are seen as maps.
Figure~\ref{fig:pred_patches} shows four of the twenty held-out test patches, chosen to span the
range of difficulty, with the AlphaEarth embedding as a principal-component false-colour, the CDL
ground truth, the random-forest prediction, and a pixel-by-pixel agreement map. On the easier
patches the prediction is almost indistinguishable from the ground truth and the field boundaries are
recovered crisply. Where errors occur they are of two kinds and both visible in the agreement panels:
small clusters of false positives where the model calls non-cultivated land cultivated, typically
along field edges and in small clearings, and false negatives on cultivated parcels the model
misses, usually narrow or fragmented fields. The errors sit on boundaries and small features rather
than over whole parcels, which is the behaviour expected from a per-pixel classifier on a
$10\,\mathrm{m}$ grid and is consistent with the cultivated class being the harder one.

%% file: sections/06_label_efficiency.tex
\section{Label Efficiency}
\label{sec:labeleff}

How much labelled data does the approach actually need? This is the practical question behind the
two training regimes of Section~\ref{sec:regimes}, and for an operational product it is often the
decisive one, since labelled pixels are the expensive ingredient.

Training on the full pool rather than the balanced sample buys surprisingly little. The random
forest improves from $92.45\%$ on 60{,}000 pixels to $93.75\%$ on the full ${\sim}8.6$ million, a
gain of about $1.3$ percentage points for roughly $140$ times as much data, and the other models
behave similarly (Table~\ref{tab:classifiers}). The balanced 60{,}000-pixel sample therefore already
captures most of the available signal, which is consistent with the rest of the analysis: the
embedding is informative enough that a small, class-balanced sample is close to sufficient, and the
expensive full-pool training adds only a modest refinement.

The comparison is in fact sharper than overall accuracy alone suggests, because the two regimes
differ in what they optimise. On the imbalance-aware metric the sampled models are the better ones:
the balanced random forest reaches $92.6\%$ balanced accuracy against $90.8\%$ for the full-pool
model, and the balanced-sample support vector machine reaches $93.0\%$ (Table~\ref{tab:classifiers}).
In other words, $140$ times more data buys about $1.3$ percentage points of overall accuracy while
costing about two points of balanced accuracy, because the extra data arrives at the natural
$78{:}22$ class ratio and pulls the decision boundary towards the majority class. For a cropland
product, where the minority cultivated class is the object of interest, the small balanced sample is
therefore not merely an acceptable compromise but arguably the better default.

Combining classifiers does not change the picture either. The ensembles, both the soft-voting
average and the stacked combiner, were trained on the same balanced 60{,}000-pixel sample as the base
models, so they are compared like for like against the sampled single classifiers. The stacked
ensemble, which learns how much to trust each base model, reaches $92.64\%$, only $0.19$ percentage
points above the balanced random forest ($92.45\%$) and still below the full-pool random forest
($93.75\%$). Learning to combine four classifiers on the small sample therefore buys less than
adding more data does, and far less than the embedding itself already provides.

Read together with the nearest-centroid baseline of Section~\ref{sec:cosine}, which reaches $90.2\%$
from two labelled class centroids and no iterative fitting, the picture is consistent: the task
is not primarily limited by the volume of labelled pixels once a balanced sample is available,
because most of the useful discriminative information is already present in the embedding
representation. This is the property that makes the paradigm attractive for operational mapping, and
it is consistent with what the model authors report for other tasks \citep{alphaearth2025,
tessera2025} and with the independent irrigated-cropland evaluation of \citet{yang2026irrigated}. The two training regimes show that a balanced sample of 60{,}000 pixels retains most of the
full-pool accuracy and achieves higher balanced accuracy at the default $0.5$ threshold. This does
not define a minimum label requirement, nor does it establish class balancing as optimal for all
applications. More broadly, the result indicates that, on agriculture-enriched held-out patches,
additional spatially correlated CDL-labelled pixels provide diminishing returns once a balanced
training sample is available.

%% file: sections/07_cross_year.tex
\section{Cross-Year Transfer}
\label{sec:crossyear}

The classification so far is built and tested within a single year, 2023. A foundation model that
produces one embedding layer per year is only useful across time if those yearly embeddings are
consistent, that is, if the 64 numbers mean the same thing in 2019 as in 2023. The cross-year
experiment tests this directly.

The design is deliberately simple. We take the random forest trained on 2023 and apply it
\emph{unchanged} to the test patches of every year from 2018 to 2023, scoring each year's
predictions against that year's CDL label. We do not retrain, fine-tune or adjust the model in any
way; only the input embeddings change. Holding the model fixed is the point of the test, not a
shortcut: it removes the classifier as a variable, so that any change in accuracy can be attributed
to the embeddings, or to the satellite inputs behind them, rather than to a fresh model fit. To read
that row correctly, however, we also need to know what each year's accuracy would be at best, so we
additionally train a separate random forest on each year and evaluate every such model on every
year. This gives a full train-year by evaluation-year matrix that separates two effects the fixed
model alone would confound: the diagonal is the per-year matched performance, showing whether a
given year is intrinsically harder to classify, while the off-diagonal entries are transfer proper.

\begin{figure}[tb]
\centering
\includegraphics[width=\linewidth]{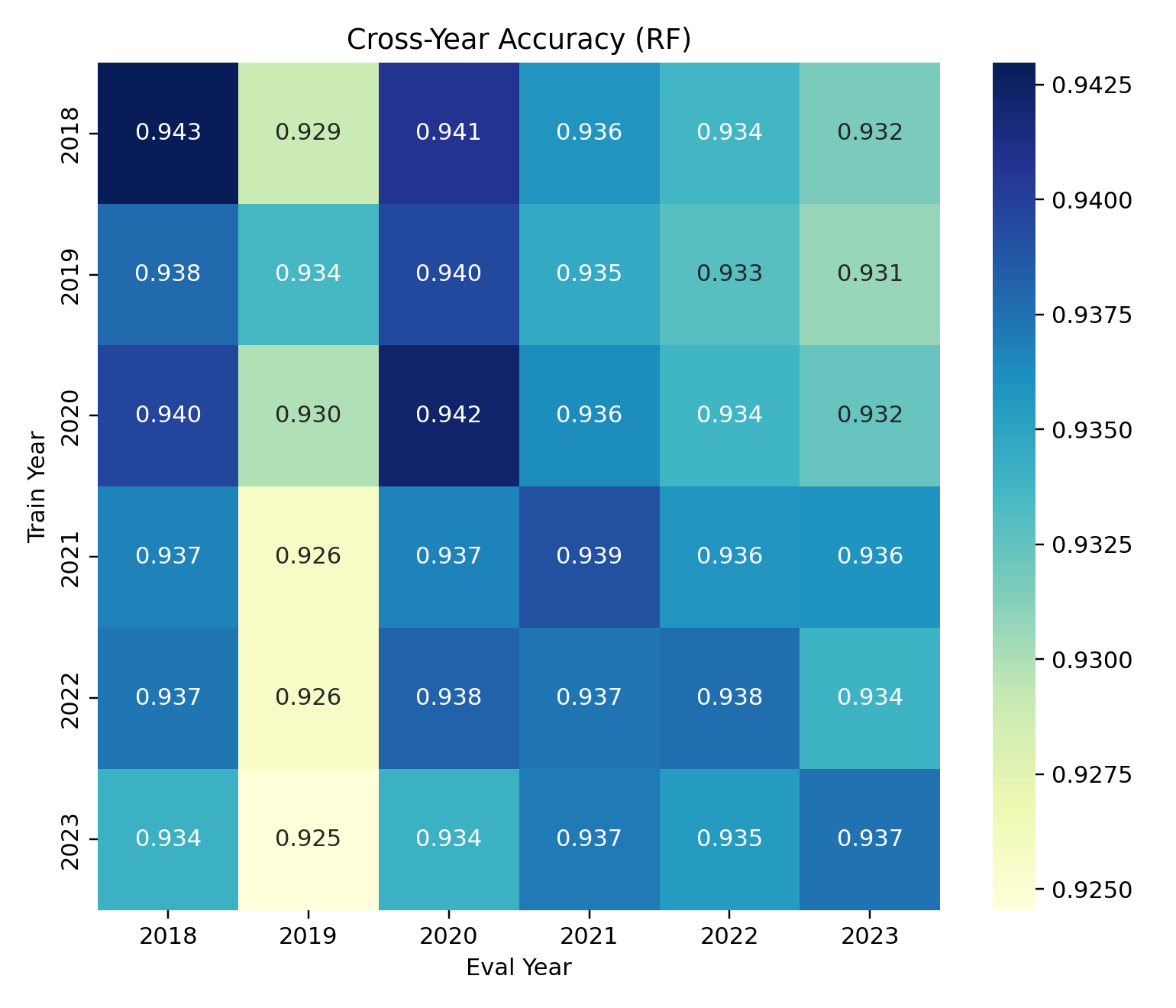}
\caption{Cross-year transfer of the random forest: overall accuracy for every combination of
training year (rows) and evaluation year (columns) on the held-out test patches. The diagonal is the
per-year matched accuracy; off-diagonal entries are temporal transfer. Accuracy is high and roughly
uniform, with 2019 the weakest year both on the diagonal and as a transfer target.}
\label{fig:crossyear}
\end{figure}

The yearly embeddings turn out to be consistent enough that a single-year model transfers with
little loss. Applying the 2023 random forest to every year gives overall accuracies between $92.5\%$
and $93.8\%$. The full matrix (Figure~\ref{fig:crossyear}) shows the same pattern across all
training years: the off-diagonal transfer entries sit close to the matched diagonal, so a model
fitted on one year and applied to another loses only a fraction of a percentage point in most cases.
The diagonal, where training and evaluation years coincide, reads $94.30\%$ (2018), $93.36\%$
(2019), $94.22\%$ (2020), $93.91\%$ (2021), $93.76\%$ (2022) and $93.75\%$ (2023).

Because overall accuracy can hide the behaviour of the minority class, we also read the diagonal on
the imbalance-aware metrics. The per-year matched balanced accuracy stays in a narrow band, $91.8\%$
(2018), $90.6\%$ (2019), $91.5\%$ (2020), $90.8\%$ (2021), $90.8\%$ (2022) and $90.8\%$ (2023), and
the cultivated-class recall likewise holds between $85.9\%$ and $87.6\%$ across the six years. The
minority class is therefore recovered consistently from year to year, not only in aggregate, and the
same holds across the full transfer matrix on balanced accuracy, where every train-evaluate
combination stays between about $86\%$ and $92\%$.

The only notable dip is 2019, which is the lowest point both on the diagonal, where it is
intrinsically the hardest year to classify, and in transfer, where the 2023 model applied to 2019
falls to about $92.5\%$. A natural guess is that a cloudier year leaves fewer clear acquisitions
behind its embeddings. We checked this against the Sentinel-2 record over the study area and do not
find support for it. The growing season of 2019 was indeed somewhat cloudier than the other years,
with fewer low-cloud scenes between April and October; however, the AlphaEarth embedding for a year
integrates all of that year's acquisitions across Sentinel-2, Landsat and Sentinel-1, so the quantity
that matters is the clear-sky availability over the full calendar year. Over the full year, 2019 is
not the cloudiest in the record: 2018 is, on every clear-sky measure, and yet 2018 is among the
better-classified years rather than the worst. Cloud cover therefore does not line up with the
accuracy pattern, so we do not attribute the small 2019 dip, a fraction of a percentage point on the
diagonal and about one point in transfer, to it, and report it as a minor difference without a single
identified cause.

The small difference between diagonal and off-diagonal performance indicates stable temporal
transfer across 2018 to 2023 within the same Maine sampling frame. However, because the same patch
locations are reused and predictions are evaluated against each year's CDL, this experiment does not
assess geographic transfer, or changes in crop calendars or reference-generation procedures. The
result should therefore be interpreted as same-region temporal transfer, rather than as general
temporal invariance of the embedding representation. It suggests that annual retraining may not be
necessary for this regional workflow, while application to new regions would still require
independent validation. We note that \citet{ma2026harvesting}, benchmarking the same
embeddings on other agricultural tasks, report limited time sensitivity as a limitation. The two
observations are plausibly the same property seen from opposite sides: annual embeddings that change
little from year to year transfer well, but are correspondingly less able to register change.

%% file: sections/08_human_validation.tex
\section{Independent Human Validation}
\label{sec:humanval}

All previous accuracy estimates use the CDL as the training and evaluation reference, not as
error-free ground truth. This section introduces an independent human-consensus reference to assess
the local agreement of the model and of the CDL on one 2023 block. The reference was built by
photo-interpretation blind to the CDL labels, and is therefore independent of the map values being
assessed, following the good-practice framework of \citet{olofsson2014good}. It then asks the
question that only becomes possible once the classifier exists: how well does the model itself agree
with the human reference?

\subsection{The continuous test block}
\label{sec:block}
The validation is carried out on the continuous test block in southern Maine introduced in
Section~\ref{sec:dataset}, an $11.2\times8.96\,\mathrm{km}$ area that is spatially separate from the
scattered training and test patches. We use a contiguous block, rather than the dispersed patches,
for two reasons: it gives a spatially coherent map that can be inspected as a whole, and it provides
a clean, independent area for the hand interpretation.

\begin{figure*}[tb]
\centering
\includegraphics[width=\textwidth]{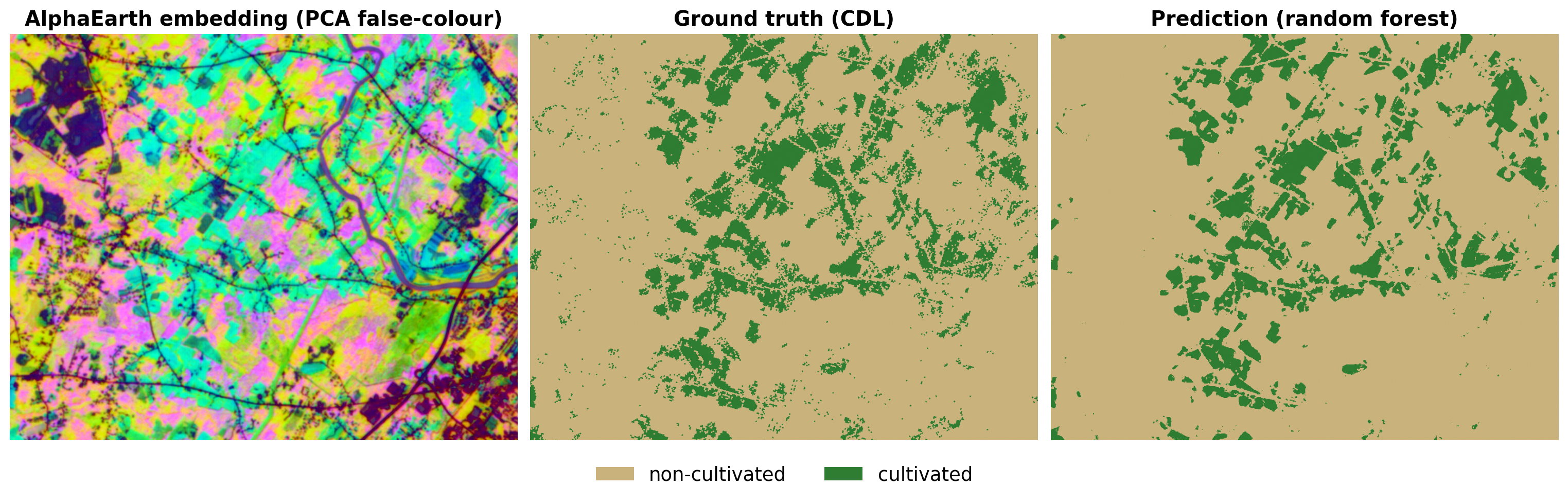}
\caption{The continuous test block (southern Maine, $11.2\times8.96\,\mathrm{km}$), 2023. Left: the
AlphaEarth embedding as a principal-component false-colour. Centre: the CDL ground truth. Right: the
random-forest prediction. The predicted cultivated pattern follows the ground truth across the whole
block (pixel-level agreement $93.3\%$).}
\label{fig:block}
\end{figure*}

Figure~\ref{fig:block} shows the whole block as the AlphaEarth false-colour, the CDL ground truth
and the random-forest prediction; the predicted cultivated pattern follows the ground truth across
the entire block. Before turning to the human reference it is worth reporting how the model does on
this block against the CDL alone, because a single contiguous landscape is a different test from an
average over twenty dispersed patches. Over the full ${\sim}1.0$ million pixels the model agrees with
the CDL on $93.3\%$ overall, but the balanced accuracy is lower, $85.7\%$, reflecting a
non-cultivated recall of $97.3\%$ against a cultivated recall of only $74.1\%$, with a cultivated
intersection-over-union of $0.655$ and a mean intersection-over-union of $0.789$. The overall
accuracy is therefore close to that of the scattered patches ($93.75\%$), but the balanced accuracy
is about five points lower than the $90.8\%$ there: a single southern-Maine block, with its own field
shapes and boundary density, is harder for the minority class than the dispersed-patch average, and
most of the difference is in cultivated recall. Because the block is a contiguous area, this is also
the only place where we report the area-based mean intersection-over-union, since it is not defined
on the scattered reference points used in the rest of this section.

\subsection{Reference sample and interpretation protocol}
\label{sec:protocol}
The number of reference points follows the standard sample-size formula \citep{cochran1977sampling},
$n_0 = z^2 p(1-p)/e^2$, with $z=1.96$ for 95\% confidence, a target margin of error $e=0.05$ and
$p=0.5$, the proportion that maximises the required sample and so the most conservative choice when
the true accuracy is unknown. This gives $n_0\approx385$, so we draw 385 points as a simple random
sample of the valid pixels of the block, with a fixed seed. We use a simple random sample rather than
a class-stratified one on purpose: stratifying by the CDL class would presuppose the very labels we
set out to test and would tie the reference design to the layer under assessment. The cost is that
the minority cultivated class receives proportionally fewer points, roughly 15\% of the sample,
which we accept so that the reference stays independent of the CDL.

Each point is interpreted in EarthLabel, a web-based photo-interpretation tool developed for this
purpose \citep{earthlabel2026}. At every point the interpreter inspects a $3\times3$ grid of
sub-points laid over very-high-resolution imagery (NAIP aerial imagery, Google satellite imagery and
Sentinel-2), which forces a judgement about the dominant land cover in the pixel's neighbourhood
rather than a single ambiguous click. The interpreted label therefore represents the dominant
land-cover condition in the immediate neighbourhood of the sampled pixel rather than an
infinitesimal point observation, which reduces sensitivity to geolocation errors and boundary
ambiguity but should be borne in mind when interpreting disagreement near field edges. For each
point the interpreter records the class, a confidence level, the agreement among the nine sub-points,
the image source used and the time taken. Crucially the interpreter works \emph{blind to the CDL
label}: if the interpreter could see the CDL value, the exercise would risk simply confirming it and
the comparison would be circular.

The 385 points were interpreted independently by two people. Before reconciliation the two agree on
$91.2\%$ of the points, with an inter-rater Cohen's kappa of $0.63$, which falls in the substantial
band of \citet{landis1977measurement}. Of the 385 points, 351 are agreed by both interpreters, and on
27 of those the agreed reading contradicts the CDL, that is, CDL errors confirmed by two
independent readings; the remaining 34 are disagreements, concentrated as expected on visually
ambiguous cases near the cultivated boundary such as fallow fields, pasture and recently disturbed
ground. These 34 were reconciled in two stages, deliberately without recourse to the CDL, which
would have reintroduced the circularity the blind protocol was meant to avoid. In the first stage 28
were settled by the confidence levels recorded at the time of interpretation, taking the more
confident reading; the confidence level is used only as this auxiliary reconciliation criterion, not
as an independent measure of whether a label is correct. The remaining 6 were genuine ties, where
both interpreters had recorded the same confidence, and were resolved in a second stage by a joint
re-examination of each point using the multi-year imagery time series, so that each point is judged
on its 2023 state, and street-view imagery where available. The result is a complete consensus
reference of 385 points, 324 non-cultivated and 61 cultivated, a $15.8\%$ cultivated share that
matches the block as a whole.

All the agreement figures below are computed on these 385 points only, since that is where the human
reference exists; they are point-level numbers and should not be confused with the block-wide,
pixel-level accuracy against the CDL reported in Section~\ref{sec:block}. From the cross-tabulation
of each map against the consensus we compute the overall agreement with a Wilson 95\% confidence
interval, the per-class user's and producer's accuracies, the balanced accuracy, and Cohen's kappa
$\kappa_{\mathrm{agr}} = (p_o-p_e)/(1-p_e)$, which corrects the raw agreement for the agreement
expected by chance.

\subsection{Is the reference layer itself trustworthy?}
\label{sec:cdltrust}
Comparing the CDL labels against the consensus reference gives an overall agreement of $91.7\%$
(95\% confidence interval $88.5\%$ to $94.1\%$) with a Cohen's kappa of $0.72$
(Table~\ref{tab:paradigm}). The agreement is high for the non-cultivated class ($F_1=0.95$) but
weaker for cultivated land ($F_1=0.77$), where the CDL shows a low user's accuracy of $69.3\%$: a
notable share of the points the CDL calls cultivated are read as non-cultivated by the interpreters.
The CDL is therefore a good but imperfect reference, with most of its error on the minority
cultivated class, which is consistent with the harder-class pattern seen throughout the
classification results.

\subsection{How well does the model agree with the human reference?}
\label{sec:modelvshuman}
The most informative comparison is between the model's own predictions and the human reference, since
it measures the quantity we ultimately care about: how well the model recovers what a human reads on
the ground, rather than how well it reproduces the CDL it was trained on. Sampling the AlphaEarth
2023 embedding at each of the 385 points and applying the chosen random forest gives an overall
agreement with the consensus of $95.3\%$ (95\% confidence interval $92.7\%$ to $97.0\%$) and a
Cohen's kappa of $0.82$.

\begin{figure}[tb]
\centering
\includegraphics[width=\linewidth]{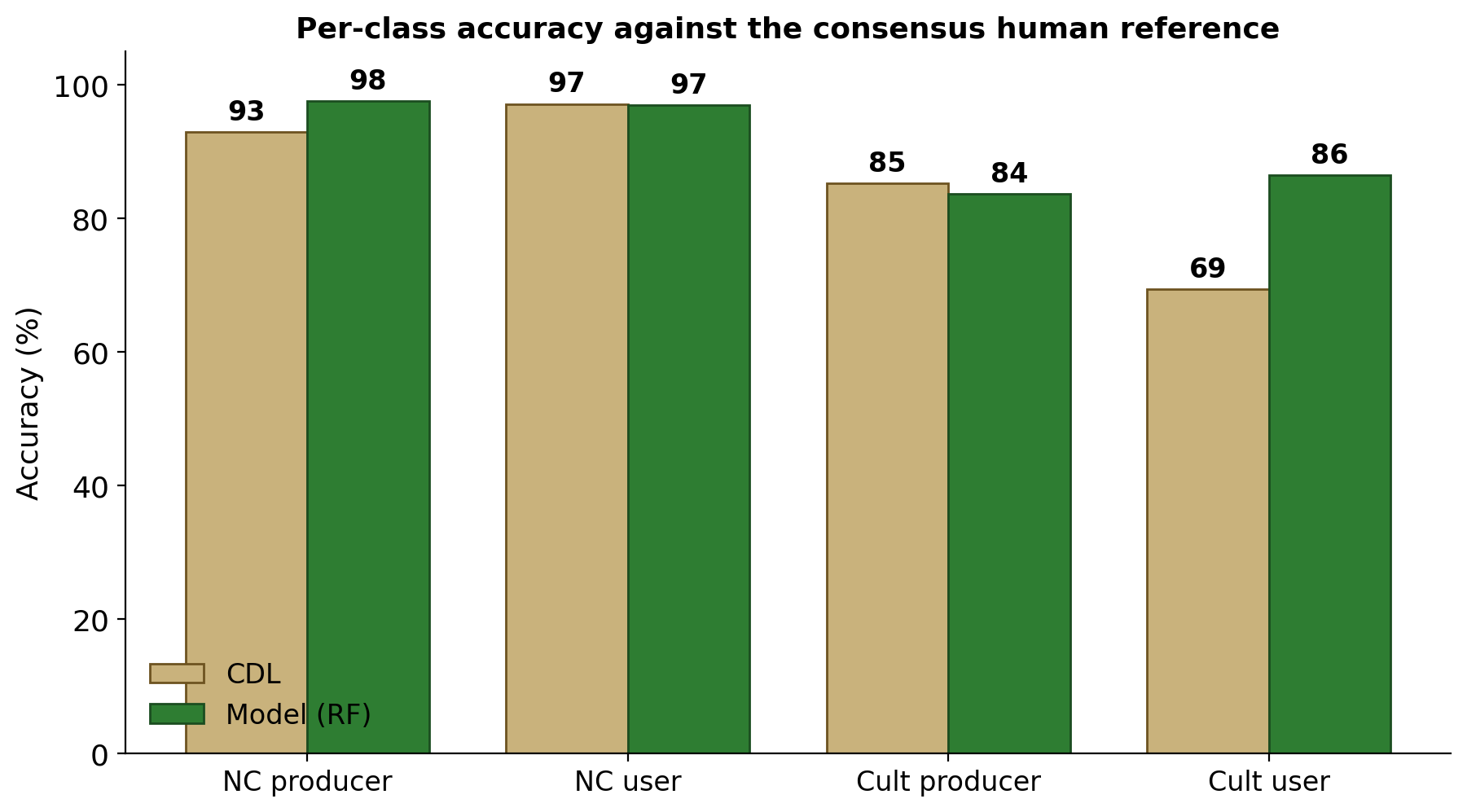}
\caption{Per-class producer's and user's accuracy against the consensus human reference, for the CDL
labels and for the random-forest predictions. The largest difference is the cultivated user's
accuracy, which rises from $69\%$ for the CDL to $86\%$ for the model: the model commits far fewer
cultivated false positives than the CDL it was trained on.}
\label{fig:pi_bars}
\end{figure}

This is the notable result of the paper: the model agrees with the human reference \emph{more
closely than the CDL labels it was trained on do}, $95.3\%$ against $91.7\%$ and $\kappa=0.82$
against $0.72$. The gain is largest exactly where the CDL is weakest, on the cultivated class, where
the model's user's accuracy is $86\%$ against the CDL's $69\%$ (Figure~\ref{fig:pi_bars}): the model
makes far fewer cultivated false positives than the raw CDL. The model appears to smooth over some
CDL pixel-level noise, in that at points where the CDL label is individually wrong the AlphaEarth
embedding may still resemble the true class, allowing the classifier to predict the human-interpreted
class correctly. In other words, training on a noisy but mostly-correct reference yields a map that
is, at these points, closer to human judgement than the reference itself.

\subsection{Comparison with a fine-tuned segmentation model}
\label{sec:paradigm}
The same consensus reference provides a fair, model-agnostic test bed on which to compare different
modelling paradigms on identical ground. As a final comparison we set the embedding-based approach
against a fine-tuned segmentation model, TerraMind \citep{terramind2025}, applied to the same
continuous test block and evaluated at the same 385 points. The two approaches are very different in
spirit: the present method places a light random forest on top of frozen AlphaEarth embeddings and
never trains a deep network, whereas TerraMind is a large multimodal backbone fine-tuned end-to-end
for the segmentation task. Holding the evaluation fixed isolates the effect of the modelling
paradigm from any difference in the test data.

Table~\ref{tab:paradigm} reports both models and the CDL baseline. The fine-tuned TerraMind reaches
$93.5\%$ overall agreement (kappa $0.75$, cultivated $F_1$ $0.79$), which places it above the CDL
baseline ($91.7\%$) but below the embedding-plus-random-forest approach ($95.3\%$). The ordering is
consistent across overall accuracy, kappa and the cultivated $F_1$. The one exception is balanced
accuracy, where TerraMind ($87.5\%$) sits just below the CDL ($89.1\%$): it agrees with the human
reference more often overall, but recovers the two classes slightly less evenly, missing a few more
of the cultivated points.

\begin{table*}[tb]
\centering
\caption{Three maps against the same consensus human reference (385 points of the continuous test
block). The confidence interval on the overall agreement is the Wilson score interval. The
area-based mean intersection-over-union is not reported here because the reference is a set of
scattered points rather than a contiguous area (Section~\ref{sec:block}).}
\label{tab:paradigm}
\small
\setlength{\tabcolsep}{5pt}
\begin{tabular}{lccccc}
\toprule
\textbf{Map (vs consensus, 385 pts)} & \textbf{OA} & \textbf{95\% CI} & \textbf{$\kappa_{\mathrm{agr}}$} & \textbf{$F_1$(cult)} & \textbf{Bal.\ acc.} \\
\midrule
\textbf{AlphaEarth $+$ random forest} & \textbf{95.3\%} & \textbf{92.7--97.0} & \textbf{0.82} & \textbf{0.85} & \textbf{90.6\%} \\
TerraMind (fine-tuned segmentation) & 93.5\% & 90.6--95.6 & 0.75 & 0.79 & 87.5\% \\
CDL (reference baseline) & 91.7\% & 88.5--94.1 & 0.72 & 0.77 & 89.1\% \\
\bottomrule
\end{tabular}
\end{table*}

\subsection{Paired significance testing}
\label{sec:mcnemar}
The accuracy figures above describe each map on its own. To ask whether one map is
\emph{significantly} closer to the human reference than another, we compare pairs of maps with
McNemar's paired test \citep{mcnemar1947note}. Because both maps in a pair are evaluated on the same
385 points, their accuracies are paired rather than independent, so the relevant question is not
simply whether one overall accuracy is higher, but whether, on the points where the two maps disagree
in correctness, one is right more often than the other. The test therefore looks only at these
discordant points and ignores the points on which both maps agree. Under the null hypothesis that the
two maps are equally accurate, each discordant point is equally likely to favour either map, so the
number won by one map follows a binomial distribution with success probability one half. We report
the exact two-sided binomial p-value, which is preferred when the discordant points are few, and read
$p<0.05$ as evidence that the difference is systematic rather than due to chance.

The improvement of the model over the CDL is statistically significant (Table~\ref{tab:mcnemar}). The
two disagree in correctness on 30 of the 385 points: on 22 the model is right where the CDL is wrong,
and on only 8 the reverse holds, which gives an exact two-sided p-value of $0.016$. On these points
the model therefore corrects significantly more CDL errors than it introduces. This does not
establish that the model is more accurate than the CDL everywhere, only that, on the 385
photo-interpreted points, it agrees with the human consensus significantly more closely than the CDL
does.

The lead over TerraMind, by contrast, is not significant. On the same points the two models disagree
in correctness on only 17: AlphaEarth with a random forest is right on 12 of these and TerraMind on
5, giving an exact two-sided p-value of $0.14$, so we cannot reject the null hypothesis that the two
are equally close to the human reference. On the 385 reference points, the accuracies of AlphaEarth with a random forest and of
TerraMind are therefore not significantly different, whereas AlphaEarth with a random forest is
significantly closer to the human consensus than the CDL. We conclude that the lightweight approach
achieves accuracy in the same observed range as TerraMind on this local reference, while avoiding
downstream fine-tuning of the AlphaEarth foundation model. For completeness, the
third pairing is also not significant: TerraMind wins 19 of its 31 discordant points against the CDL
and the CDL 12, an exact two-sided $p=0.28$, even though its overall agreement is higher. Among the
three maps, then, only the embedding-based model significantly outperforms the CDL reference. All
three tests use the exact binomial formulation on the paired predictions over the common
photo-interpreted set, at the $\alpha=0.05$ level.

\begin{table*}[tb]
\centering
\caption{McNemar paired comparisons against the human consensus reference, on the 385 points. Only
the discordant points, where exactly one of the two maps is correct, enter each test; the concordant
points carry no information about which map is better. ``A wins'' counts the discordant points on
which map A matches the human reference and map B does not. Only the model against the CDL is
significant at the 5\% level.}
\label{tab:mcnemar}
\small
\setlength{\tabcolsep}{6pt}
\begin{tabular}{lccccl}
\toprule
\textbf{Pair (A vs B)} & \textbf{A wins} & \textbf{B wins} & \textbf{Discordant} & \textbf{Exact $p$} & \textbf{Verdict} \\
\midrule
AlphaEarth $+$ RF vs CDL & 22 & 8 & 30 & \textbf{0.016} & significant \\
AlphaEarth $+$ RF vs TerraMind & 12 & 5 & 17 & 0.14 & not significant \\
TerraMind vs CDL & 19 & 12 & 31 & 0.28 & not significant \\
\bottomrule
\end{tabular}
\end{table*}

This is a single comparison on one test block, interpreted by the same two people, so it should be
read as indicative rather than as a definitive ranking of the two paradigms. What it does establish
is that a light classifier on frozen embeddings is not obviously behind a large model fine-tuned
end-to-end for the task, at a small fraction of the training cost.

%% file: sections/09_discussion.tex
\section{Discussion}
\label{sec:discussion}

\subsection{Why a classifier can be closer to the truth than its own training labels}
The result that most needs explaining is the one in Section~\ref{sec:modelvshuman}: a model trained
on the CDL agrees with independent human interpretation more closely than the CDL does. This sounds
paradoxical only if one expects a classifier to reproduce its labels. What it actually learns is the
regularity in the relationship between embedding and label, and a per-pixel error in the reference
is, by construction, not a regularity. Where the CDL misassigns an isolated pixel, that pixel's
embedding still resembles the class it truly belongs to, because the embedding is computed from
imagery and not from the label; the classifier, having fitted the dominant pattern across 8.6 million
pixels, therefore predicts the class the embedding supports rather than the erroneous label. Label
noise that is unsystematic averages out during fitting, and the fitted model is smoother than the
reference it was fitted to. The effect is visible in exactly the place the mechanism predicts: the
CDL's weakest quantity is its cultivated user's accuracy ($69.3\%$), that is, cultivated commission
errors, and this is where the model gains most ($86\%$).

Two qualifications keep this from being over-read. The argument holds for noise that is not
systematically aligned with the embedding: a bias the CDL applies consistently to a whole cover type
would be learned rather than smoothed away, and our data cannot distinguish the two cases. And the
evidence is 385 points on one block, which is enough for the paired test to reach significance
against the CDL ($p=0.016$) but not enough to characterise CDL error across Maine or across years.

\subsection{Operational implications}
Read together, the results describe a low-compute prototype workflow rather than a
deployment-ready statewide product. The user does not train or fine-tune the foundation model; the
principal downstream cost is fitting a standard classifier to downloaded embeddings. The workflow is
reproducible, and a balanced pixel sample captures most of the discrimination observed on the
held-out patches. Operational use would in any case require validation on a probability-based sample
of the target area, calibration that accounts for class prevalence, area-adjusted accuracy and
uncertainty estimates, monitoring for year-to-year distribution shifts, and evaluation across
different agricultural settings. Nevertheless, the present study provides a regional baseline from
which these operational requirements can be developed.

The limits of that claim should be stated as plainly. The study covers one state whose land cover is
dominated by forest and where cultivated land is sparse and spatially fragmented, so the numbers may
not transfer unchanged to regions with larger continuous agricultural areas, different crop
calendars, or more heterogeneous farming systems; the independent evaluation of
\citet{yang2026irrigated} reports exactly this pattern, with stable transfer between years but
weaker generalisation across regions, and \citet{ma2026harvesting} likewise find limited
spatial transferability of these embeddings relative to purpose-built remote-sensing models. Their
strongest case is worth stating plainly: transferring a county-level soybean yield model from the
United States to Argentina, embedding-based models return a negative coefficient of determination in
every year from 2019 to 2024, between $-0.64$ and $-5.24$, against $0.27$ overall for the
remote-sensing baseline. Transfer to a genuinely different agricultural setting is not a matter of
some lost accuracy but of a model that may fail outright, which is the risk our own single-state
scope leaves untested. The
classification is also purely per-pixel, so the residual
errors concentrate at field boundaries and on small or fragmented parcels rather than over whole
fields, and the balanced accuracy on the contiguous block ($85.7\%$) is about five points below the
dispersed-patch average, almost entirely through cultivated recall. A product built this way would
inherit that behaviour.

\subsection{When frozen embeddings may be preferable to fine-tuning}
The comparison of Section~\ref{sec:paradigm} does not show that frozen embeddings beat fine-tuning:
the difference against TerraMind is not statistically significant ($p=0.14$), and on a single block
interpreted by two people it could not carry that weight even if it were. What it does suggest is
that the two are in the same range on this task, which shifts the question from accuracy to cost.
On that axis the frozen route is clearly cheaper, and the conditions under which it appears
preferable are visible in our own results: when the task is a coarse, binary distinction that the
representation already encodes, as the $90.2\%$ nearest-centroid baseline of Section~\ref{sec:cosine}
indicates; when labels are limited, since a balanced sample of 60{,}000 pixels is close to
sufficient; when a multi-year product is wanted from a single fitting; and when there is no capacity
to train or serve a deep network. Conversely, the cases where fine-tuning should still be expected
to pay are those our design cannot address: tasks needing finer distinctions than the embedding was
trained to preserve, tasks where field-level shape and boundary quality matter more than per-pixel
agreement, given where our residual errors sit, and settings where a modality the embedding does not
carry is decisive.

We note finally that the choice is not only technical. AlphaEarth is released as embeddings only,
with the model itself closed, so fine-tuning it is not an option available to a user in any case;
TESSERA releases both. Which of the two families a practitioner can adopt is therefore partly a
question of what has been made public.

\subsection{Limitations and scope of inference}
\label{sec:limitations}
Some limitations have to be considered in analysing these results. First, the study is
restricted to Maine, a forest-dominated state with sparse and fragmented cultivation, and the
agriculture-enriched patch sample is not intended to be representative of the state as a whole.
Second, the reference labels originate from a $30\,\mathrm{m}$ classified product aligned to a
$10\,\mathrm{m}$ embedding grid, which limits the interpretation of pixel-level accuracy,
particularly near class boundaries. Spatial autocorrelation among neighbouring pixels further
reduces the effective number of independent observations, although the spatial separation of the
test patches limits direct train-test leakage. The temporal experiment evaluates transfer across
years at the same locations within Maine, and should not be interpreted as evidence of geographic
generalisation. Finally, even though this is quite common in validation, the independent human
reference consists of 385 points, of which 61 are cultivated, based on image interpretation rather
than field observation. These limitations constrain the generality of the findings, but they do not
affect the central result that simple classifiers can achieve strong predictive performance from
frozen AlphaEarth embeddings within the evaluated setting.

%% file: sections/10_conclusions.tex
\section{Conclusions}
\label{sec:conclusions}

We set out to test whether embedding-based geospatial foundation models, and AlphaEarth in
particular, can deliver accurate and label-efficient cropland classification without training or
fine-tuning a deep network. On a binary, per-pixel classification of cultivated against
non-cultivated land over Maine, with labels derived from the USDA Cropland Data Layer and the annual
embeddings as the only input, a lightweight classifier on the frozen 64-dimensional vectors reaches
about $93.7\%$ overall accuracy and $90.8\%$ balanced accuracy on held-out patches, with a
macro-averaged $F_1$ of $0.898$.

The choice of classifier matters little once the embedding is fixed: logistic regression
($93.48\%$) sits within $0.3$ percentage points of a gradient-boosted ensemble ($93.75\%$), and a
nearest-class-centroid rule, which estimates one labelled centroid per class and fits nothing
further, already reaches $90.2\%$. Nor is the task primarily limited by the volume of labelled
pixels: a balanced sample of 60{,}000 recovers most of the accuracy of the full pool of roughly 8.6
million, and does so at a higher balanced accuracy. The statistical characterisation accounts for
both findings. The embeddings lie on the unit hypersphere and form tightly concentrated angular
clusters; every dimension departs from normality, with mild skew rather than heavy tails the
dominant feature, so classical Gaussian coordinate-wise tools are not the appropriate description;
and the class signal is spread thinly across many dimensions rather than held in a few. That is why
a linear boundary is nearly sufficient, and why, under a criterion of no measurable loss, no
dimension can be discarded for free.

Two further results complete the picture. The yearly embeddings are consistent enough that a
classifier fitted on 2023 transfers to every year from 2018 to 2023, within the same Maine sampling
frame, with little loss. And when the CDL is set aside in favour of an independent two-interpreter
consensus on a contiguous block, the model agrees with the human reference more closely ($95.3\%$,
$\kappa=0.82$) than the CDL it was trained on does ($91.7\%$, $\kappa=0.72$), a difference that is
significant on the paired points ($p=0.0161$), while remaining statistically indistinguishable from
a fine-tuned segmentation model ($93.5\%$, $p=0.14$). Within the limits set out in
Section~\ref{sec:limitations}, frozen geospatial embeddings are a competitive and inexpensive basis
for regional cropland mapping.

The most immediate extensions follow from those limits: a like-for-like comparison with another
openly released embedding model such as TESSERA on the same patches and labels
\citep{lisaius2026senegal}; a broader comparison against fine-tuned segmentation, over contiguous
tiles rather than scattered points and retaining the paired testing used here; and the addition of
light spatial context to the per-pixel classifier, since the residual error concentrates at field
boundaries and on small parcels. Testing the same workflow across more varied agricultural regions
remains the clearest route to establishing how far a single representation travels.